\documentclass[journal]{IEEEtran}
\usepackage{times}
\usepackage{epsfig}
\usepackage{graphicx}
\usepackage{amsmath}
\usepackage{amssymb}
\usepackage{amsthm,thmtools}
\usepackage{cite}
\usepackage{booktabs}
\usepackage{algorithmic}
\usepackage{algorithm}
\usepackage{tabularx}
\usepackage{longtable}
\usepackage{bm}
\usepackage{color}
\usepackage[pagebackref=true,breaklinks=true,letterpaper=true,colorlinks,citecolor=blue,linkcolor=blue,bookmarks=false]{hyperref}
\usepackage{url}
\usepackage{caption}
\newcommand{\ignore}[1]{}

\usepackage{graphicx}
\usepackage{algorithm}
\usepackage{amsmath,amssymb} 
\usepackage{color}

\declaretheoremstyle[
  spaceabove=6pt plus 6pt,
  spacebelow=6pt plus 6pt,
  headfont=\itshape,
  bodyfont=\normalfont,
  postheadspace={ },
  qed=\protect\thisproofqed,
]{proof}

\declaretheorem[
  name=\protect\thisproofname,
  style=proof,
  numbered=no,
]{gproof}

\newcommand\thisproofname{}
\newcommand\thisproofqed{}

\newenvironment{proof*}[1][\proofname]
  {\renewcommand\thisproofname{#1}%
   \renewcommand\thisproofqed{\openbox}%
   \gproof}
  {\endgproof}

\ifCLASSOPTIONcompsoc
\else
\fi

\begin{document}
%
\title{Visual Tracking via Boolean Map Representations}

\author{{\quad\\}Kaihua~Zhang, Qingshan~Liu, 
       and Ming-Hsuan~Yang
\IEEEcompsocitemizethanks{
\IEEEcompsocthanksitem Kaihua~Zhang and Qingshan~Liu are with Jiangsu Key Laboratory of Big Data Analysis Technology (B-DAT), Nanjing University of Information Science and Technology.
E-mail: \{cskhzhang, qsliu\}@nuist.edu.cn.
\IEEEcompsocthanksitem Ming-Hsuan Yang is with Electrical Engineering and Computer Science, University of California, Merced, CA, 95344. E-mail: mhyang@ucmerced.edu.}
\thanks{}}

\markboth{Submitted}%
{Shell \MakeLowercase{\textit{et al.}}: Bare Demo of IEEEtran.cls for Computer Society Journals}
\IEEEcompsoctitleabstractindextext{%
\begin{abstract}
In this paper, we present a simple yet effective Boolean map based representation
that exploits connectivity cues for visual tracking.
We describe a target object with histogram of oriented gradients
and raw color features, of which each one
is characterized by a set of Boolean maps generated by uniformly thresholding their values.
The Boolean maps effectively encode multi-scale connectivity cues of the target with different granularities.
The fine-grained Boolean maps capture spatially structural details that are effective
for precise target localization
while the coarse-grained ones encode global shape information
that are robust to large target appearance variations.
Finally, all the Boolean maps form together a robust representation that can be approximated
by an explicit feature map of the intersection kernel,
which is fed into a logistic regression classifier with online update, and
the target location is estimated within a particle filter framework.
The proposed representation scheme is computationally efficient and
facilitates achieving favorable performance in terms of accuracy and robustness
against the state-of-the-art tracking methods on a large benchmark dataset of 50 image sequences.
\end{abstract}

}

\maketitle

\IEEEdisplaynotcompsoctitleabstractindextext

%
\IEEEpeerreviewmaketitle

\begin{figure*}[tb]
\begin{center}
 \includegraphics[width=.96\linewidth]{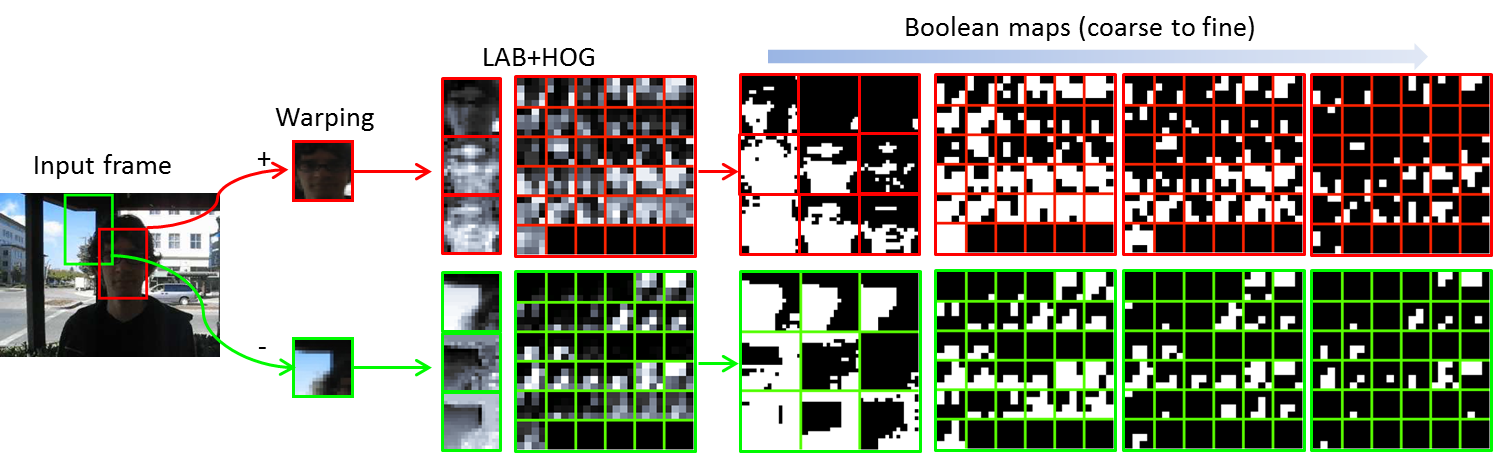}
\end{center}
   \caption{Boolean map representation. For clarity, only the BMRs of a positive and negative sample are demonstrated.
Note that the Boolean maps contain more connected structures than the LAB+HOG representations.}
\label{fig:principle}
\end{figure*}
\section{Introduction}

Object tracking is a fundamental problem in computer vision and image processing
with numerous applications. Despite significant progress in past decades, it remains a challenging task due to large appearance variations caused by illumination
changes, partial occlusion, deformation, as well as cluttered background.
To address these challenges, a robust representation plays a critical
role for the success of a visual tracker, and attracts much attention in recent years~\cite{li2013survey}.
%

%
%
Numerous representation schemes have been developed for visual tracking based on
holistic and local features.
Lucas and Kanade~\cite{lucas1981iterative} leverage holistic templates based on raw pixel values
to represent target appearance.
Matthews et al.~\cite{matthews2004template} design an effective template update scheme that
uses stable information from the first frame for visual tracking.
In~\cite{henriques2012circulant} Henriques et al. propose a correlation filter based template (trained with raw intensity) for visual tracking with promising performance.
Zhang et al.~\cite{zhang2014meem} propose a multi-expert restoration scheme to address the drift problem in tracking, in which each base tracker leverages an explicit feature map representation via quantizing the CIE LAB color channels of spatially sampled image patches.
To deal with appearance changes, subspace learning based trackers have been proposed.
Black and Jepson~\cite{black1998eigentracking} develop a pre-learned view-based eigenbasis representation for visual tracking.
However, the pre-trained representation cannot adapt well to significant target appearance variations.
In~\cite{Ross_IJCV_2008} Ross et al. propose an incremental update scheme to learn a low-dimensional subspace representation.
Recently, numerous tracking algorithms based on sparse representation have been proposed. Mei and Ling~\cite{Mei_PAMI_2011} devise a dictionary of holistic intensity templates with target and trivial templates, and then find the location of the object with minimal reconstruction error via solving an $\ell_1$ minimization problem.
Zhang et al.~\cite{zhang2012robust} formulate visual tracking as a multi-task sparse learning problem, which learns particle representations jointly.
In~\cite{wang2013online} Wang et al. introduce $\ell_1$ regularization into the eigen-reconstruction to develop an effective representation that combines the merits of both subspace and sparse representations.

In spite of demonstrated success of exploiting global representations for visual tracking, existing methods
are less effective in dealing with heavy occlusion and large deformation as local visual cues
are not taken into account.
Consequently, local representations are developed to handle occlusion and deformation.
Adam et al.~\cite{adam2006robust} propose a fragment-based tracking method that divides a target object into a set of local regions and represents each region with a histogram.
In~\cite{he2013visual}, He et al. present a locality sensitive histogram for visual tracking by
considering the contributions of local regions at each pixel, which can model target appearance well.
Babenko et al.~\cite{Babenko_PAMI_2011} formulate the tracking task as a multiple instance learning problem, in which Haar-like features are used to represent target appearance.
Hare et al.~\cite{Hare_ICCV_2011} pose visual tracking as a structure learning task and
leverage Haar-like features to describe target appearance.
In~\cite{henriques2015high} Henriques et al. propose an algorithm based on
a kernel correlation filter (KCF) to describe target templates with
feature maps based on histogram of oriented gradients (HOG)~\cite{dalal2005histograms}.
This method has been shown to achieve
promising performance on the recent tracking benchmark dataset~\cite{wu2013online}
in terms of accuracy and efficiency.
Kwon and Lee~\cite{kwon2009tracking} present a tracking method that represents
target appearance with a set of local patches where the topology is updated to account for
large shape deformation.
Jia et al.~\cite{jia2012visual} propose a structural sparse representation scheme
by dividing a target object into some local image patches in a regular grid and using the coefficients
to analyze occlusion and deformation.

Hierarchical representation methods that capture holistic and local object appearance
have been developed for visual tracking \cite{zhong2012robust,li2014scale,wang2015understanding,ma2015hierarchical}.
%
%
Zhong et al.~\cite{zhong2012robust} propose a sparse collaborative appearance model
for visual tracking in which both holistic templates and local representations are used.
Li and Zhu~\cite{li2014scale} extend the KCF tracker~\cite{henriques2015high} with
a scale adaptive scheme and effective color features.
In~\cite{wang2015understanding},
Wang et al. demonstrate that a simple tracker based on logistic regression with a
representation composed of HOG and raw color channels performs favorably on the
benchmark dataset~\cite{wu2013online}.
Ma et al.~\cite{ma2015hierarchical} exploit features from hierarchical
layers of a convolutional neural network
and learn an effective KCF which takes account of spatial details and semantics of target objects
for visual tracking.
%
%

In biological vision, it has been suggested that object tracking is carried out by
attention mechanisms~\cite{allen2004attention,cavanagh2005tracking}.
%
Global topological structure such as connectivity is used to model tasks
related to visual attention~\cite{set1982topological,palmer1999vision}.
However, all aforementioned representations do not consider
topological structure for visual tracking.

In this work, we propose a Boolean map based representation (BMR) that leverages
connectivity cues for visual tracking.
One case of connectivity is the enclosure topological relationship
between the (foreground) figure and ground which defines the boundaries of figures.
Recent gestalt psychological studies suggest that the enclosure topological cues play an important role in
figure-ground segregation and have been successfully applied to
saliency detection~\cite{zhang2013saliency} and measuring
objectness~\cite{alexe2012measuring,cheng2014bing}.
The proposed BMR scheme characterizes target appearance by concatenating multiple layers of Boolean maps at different granularities based on uniformly thresholding HOG and color feature maps.
The fine-grained Boolean maps capture locally spatial structural details that are effective for precise localization and coarse-grained ones which encode much global shape information to account for
significant appearance variations. 
The Boolean maps are then concatenated and normalized to a BMR scheme
that can be just approximated by an explicate feature map.
We learn a logistic regression classifier using the BMR scheme and online update to estimate target locations within a particle filter framework.
The effectiveness of the proposed algorithm is demonstrated on a large tracking benchmark dataset
with 50 challenging videos~\cite{wu2013online} against the state-of-the-art approaches.

The main contributions of this work are summarized as follows:
\begin{itemize}
\item We demonstrate that the connectivity cues can be effectively used for robust visual tracking.
  \item We show that the BMR scheme can be approximated as an explicit feature map of
the intersection kernel which can find a nonlinear classification boundary via a linear classifier. In addition,
it is easy to train and detect for robust visual tracking with this approach.
  \item The proposed tracking algorithm based on the BMR scheme performs favorably
in terms of accuracy and robustness to initializations based on the benchmark dataset with 50 challenging videos~\cite{wu2013online} against 35 methods
including the state-of-the-art trackers based on hierarchical features from deep
networks~\cite{ma2015hierarchical} and multiple experts with entropy minimization
(MEEM)~\cite{zhang2014meem}.
\end{itemize}

\section{Tracking via Boolean Map Representations}
We present the BMR scheme and a logistic regression classifier
with online update for visual tracking.

\subsection{Boolean Map Representation}
The proposed image representation is
based on recent findings of human visual attention~\cite{huang2007boolean} which shows that momentary conscious awareness of a scene can
be represented by Boolean maps.
The Boolean maps are concerned with center-surround contrast that mimic the sensitivity of neurons either to dark centers on bright surrounds or vice versa~\cite{itti1998model}.
Specifically, we exploit the connectivity cues inside a target measured by the Boolean maps
which can be used for separating the foreground object
from the background effectively~\cite{set1982topological,zhang2013saliency,alexe2012measuring,cheng2014bing}.
As demonstrated in Figure~\ref{fig:principle}, the connectivity inside a target can be well captured by the Boolean maps at different scales.

Neurobiological studies have demonstrated that human visual system is sensitive to color and edge orientations~\cite{livingstone1984anatomy} which provide
useful cues to discriminate the foreground object from the background.
In this work, we use color features in the CIE LAB color space and HOG features to represent objects.
To extract the perceptually uniform color features, we first normalize each sample $\mathbf{x}$ to a canonical size ($32\times 32$ in our experiments),
and then subsample it to a half size to reduce appearance variations, and finally transform the sample into the CIE LAB color space, denoted as $\bm{\Phi}^{col}(\mathbf{x})\in \mathbb{R}^{n_{col}\times n_{col}\times 3}$ ($n_{col}=16$ in this work).
%
%
Furthermore, we leverage the HOG features to capture edge orientation information of a target object,
denoted as $\bm{\Phi}^{hog}(\mathbf{x})\in \mathbb{R}^{n_{hog}\times n_{hog}\times 31}$ ($n_{hog}=4$ in this work).
%
Figure~\ref{fig:principle} demonstrates that most color and HOG feature maps of the target own center-surrounded patterns that are similar to biologically plausible architecture of primates in~\cite{itti1998model}.
We normalize both $\bm{\Phi}^{col}(\mathbf{x})$ and $\bm{\Phi}^{hog}(\mathbf{x})$ to range from 0 to 1, and
concatenate $\bm{\Phi}^{col}(\mathbf{x})$ and $\bm{\Phi}^{hog}(\mathbf{x})$ to form a feature vector $\bm{\phi}(\mathbf{x})\in \mathbb{R}^{d\times 1}$
with $d=3n_{col}^2+31n_{hog}^2$.
The feature vector is rescaled  to $[0,1]$ by
\begin{equation}
\bm\phi(\mathbf{x})\leftarrow\frac{\bm\phi(\mathbf{x})-\min(\bm\phi(\mathbf{x}))}{\max(\bm\phi(\mathbf{x}))-\min(\bm\phi(\mathbf{x}))},
\label{eq:scalephi}
\end{equation}
where $\max(\cdot)$ and $\min(\cdot)$ denotes the maximal and minimal operators, respectively.

Next, $\bm\phi(\mathbf{x})$ in (\ref{eq:scalephi}) is encoded into a set of vectorized Boolean maps $\mathcal{B}(\mathbf{x})=\{\mathbf{b}_i(\mathbf{x})\}_{i=1}^c$
by
\begin{equation}
\mathbf{b}_i(\mathbf{x})=
\left\{
\begin{aligned}
1 & , & \bm{\phi}(\mathbf{x})\succeq \theta_i,\\
0 & , & \mbox{otherwise},\\
\end{aligned}
\right.
\label{eq:booleanMap}
\end{equation}
where $\theta_i\thicksim U(0,1)$ is a threshold drawn from a uniform distribution over $[0,1]$, and the symbol $\succeq$ denotes elementwise inequality.
In this work, we set $\theta_i=i/c$ that is simply sampled at a fixed-step size $\delta=1/c$, and a fixed-step sampling is equivalent to the uniform sampling
in the limit $\delta\rightarrow 0$~\cite{zhang2013saliency}.
Hence, we have $\mathbf{b}_1(\mathbf{x})\succeq\mathbf{b}_2(\mathbf{x})\succeq\ldots\succeq\mathbf{b}_c(\mathbf{x})$. It is easy to show that
\begin{figure}[t]
\begin{center}
 \includegraphics[width=.95\linewidth]{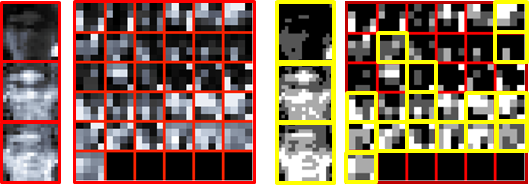}
\end{center}
   \caption{Right two columns: reconstructed LAB+HOG representations of the target by BMRs in our experiments. Left two columns: the corresponding prototypes shown in Figure~\ref{fig:principle}. Some reconstructed ones with more connected structures than their prototypes are highlighted in yellow.}
\label{fig:reconstruction}
\end{figure}
\begin{equation}
0\leq\phi_k(\mathbf{x})-\frac{1}{c}\sum_{j=1}^cb_{j,k}(\mathbf{x})<\delta,
\label{eq:phibound}
\end{equation}
where $\phi_k$ and $b_{j,k}$ are the $k$-th entries of $\bm\phi$ and $\mathbf{b}_j$, respectively.

\begin{proof*}
Without loss of generality, we assume that $i\delta\leq\phi_k(\mathbf{x})<(1+i)\delta$, $i=0,\ldots,c$.
As such, we have $b_{j,k}(\mathbf{x})=1$ for all $j\leq i$ because $\mathbf{b}_1(\mathbf{x})\succeq\mathbf{b}_2(\mathbf{x})\succeq\ldots\succeq\mathbf{b}_c(\mathbf{x})$, and $b_{j,k}(\mathbf{x})=0$ for $j>i$.
Therefore, we have $\frac{1}{c}\sum_{j=1}^cb_{j,k}(\mathbf{x})=i\delta$, and $0\leq\phi_k(\mathbf{x})-\frac{1}{c}\sum_{j=1}^cb_{j,k}(\mathbf{x})<(1+i)\delta-i\delta=\delta $.
\end{proof*}

In (\ref{eq:phibound}), when $\delta\rightarrow 0$ (i.e., $\theta_i\thicksim U(0,1)$) we have
\begin{equation}
\phi_k(\mathbf{x})=\frac{1}{c}\sum_{j=1}^cb_{j,k}(\mathbf{x}).
\label{eq:phkeqbik}
\end{equation}

In this work, we set $\delta=0.25$.
Although (\ref{eq:phkeqbik}) may not be strictly satisfied, empirical results show that most distinct structures in $\bm\phi(\mathbf{x})$ can be
reconstructed as demonstrated in Figure~\ref{fig:reconstruction}.
%
Furthermore, the reconstructed representations contain more connected structures than the original ones (see the ones highlighted in yellow in Figure~\ref{fig:reconstruction}), which
shows that the Boolean maps facilitate capturing global geometric information of target objects.

Based on (\ref{eq:phkeqbik}), to measure the similarity between two samples $\mathbf{x}$ and $\mathbf{y}$, we use
the intersection function~\cite{grauman2007pyramid}
\begin{equation}
\begin{aligned}
\widetilde{\mathcal{I}}(\bm{\phi}(\mathbf{x}),\bm{\phi}(\mathbf{y}))&=\sum_{k=1}^d\min(\phi_k(\mathbf{x}),\phi_k(\mathbf{y}))\\
&=\sum_{k=1}^d\min(\frac{1}{c}\sum_{j=1}^cb_{j,k}(\mathbf{x}), \frac{1}{c}\sum_{j=1}^cb_{j,k}(\mathbf{y}))\\
&=\frac{1}{c}\sum_{k=1}^d\sum_{j=1}^c\min( b_{j,k}(\mathbf{x}), b_{j,k}(\mathbf{y}))\\
&=\frac{1}{c}\sum_{k=1}^d\sum_{j=1}^c b_{j,k}(\mathbf{x})b_{j,k}(\mathbf{y})\\
&=<\widetilde{\mathbf{b}}(\mathbf{x}),\widetilde{\mathbf{b}}(\mathbf{y})>,
\end{aligned}
\label{eq:minikernel}
\end{equation}
where $\widetilde{\mathbf{b}}=[\mathbf{b}_1^\top,\ldots,\mathbf{b}_c^\top]^\top/\sqrt{c}$.

To avoid favoring larger input sets~\cite{grauman2007pyramid}, we normalize $\widetilde{\mathcal{I}}(\bm{\phi}(\mathbf{x}),\bm{\phi}(\mathbf{y}))$ in (\ref{eq:minikernel})
and define the kernel $\mathcal{I}(\bm{\phi}(\mathbf{x}),\bm{\phi}(\mathbf{y}))$ as
\begin{equation}
\begin{aligned}
\mathcal{I}(\bm{\phi}(\mathbf{x}),\bm{\phi}(\mathbf{y}))&=\frac{\widetilde{\mathcal{I}}(\bm{\phi}(\mathbf{x}),\bm{\phi}(\mathbf{y}))}{\sqrt{\widetilde{\mathcal{I}}(\bm{\phi}(\mathbf{x}),\bm{\phi}(\mathbf{x}))\widetilde{\mathcal{I}}(\bm{\phi}(\mathbf{y}),\bm{\phi}(\mathbf{y}))}}\\
&=<\mathbf{b}(\mathbf{x}),\mathbf{b}(\mathbf{y})>,
\label{eq:normalizekernel}
\end{aligned}
\end{equation}
where $\mathbf{b}(\cdot)$ is an explicit feature map function.
In this work, the feature map function is defined by
\begin{equation}
\mathbf{b}(\mathbf{x}) = \frac{\widetilde{\mathbf{b}}(\mathbf{x})}{|\widetilde{\mathbf{b}}(\mathbf{x})|_2^1},
\label{eq:explicitmap}
\end{equation}
where $|\cdot|_2^1$ is an $\ell_2$ norm operator.
We use $\mathbf{b}(\mathbf{x})$ to train a linear classifier, which is able to address the nonlinear classification problem in the feature space of $\bm{\phi}$ for visual tracking with favorable performance.
The proposed tracking algorithm based on BMR is summarized in Algorithm~\ref{alg:BMR}.
\begin{algorithm}[t]
\caption{BMR}
\begin{algorithmic}\label{alg:BMR}
\STATE \textbf{Input:} Normalized image patch $\mathbf{x}$;
\begin{enumerate}\setlength{\itemsep}{-\itemsep}
\item Compute feature vector $\bm\phi(\mathbf{x})$ in (\ref{eq:scalephi});
\item \textbf{for all} entries $\phi_i(\mathbf{x}), i=1,\ldots,d$ of $\bm\phi(\mathbf{x})$, \textbf{do}
\item\quad \textbf{for} $\theta_i=\delta:\delta:1-\delta$, \textbf{do}
\item\quad\quad \textbf{if} $\phi_i(\mathbf{x})>\theta_i$;
\item\quad\quad\quad$b_i(\mathbf{x})=1$;
\item\quad\quad \textbf{else}
\item\quad\quad\quad$b_i(\mathbf{x})=0$;
\item\quad\quad \textbf{end if}
\item\quad\textbf{end for}
\item\textbf{end for}
\item $\widetilde{\mathbf{b}}(\mathbf{x})\leftarrow[b_1(\mathbf{x}),\ldots,b_{cd}(\mathbf{x})]^\top/\sqrt{c}$;
\item $\mathbf{b}(\mathbf{x})\leftarrow \widetilde{\mathbf{b}}(\mathbf{x})/|\widetilde{\mathbf{b}}(\mathbf{x})|_2^1$ $//\textit{Normalization}$;
\end{enumerate}
\STATE \textbf{Output:} BMR $\mathbf{b}(\mathbf{x})$.
\end{algorithmic}
\end{algorithm}

\subsection{Learning Linear Classifier with BMRs}
We pose visual tracking as a binary classification problem with local search, in which a linear classifier is learned in the Boolean map feature space to separate
the target from the background.
Specifically, we use a logistic regressor to learn the classifier for measuring similarity scores of samples.

Let $\mathbf{l}_t(\mathbf{x}_t^i)\in \mathbb{R}^2$ denote the location of the $i$-th sample at frame $t$.
We assume that $\mathbf{l}_t(\hat{\mathbf{x}}_t)$ is the object location, and densely draw samples $\mathcal{D}^\alpha=\{\mathbf{x}|||\mathbf{l}_t(\mathbf{x})-\mathbf{l}_t(\hat{\mathbf{x}}_t)||<\alpha\}$ within a search radius $\alpha$ centered at the current object location, and label them as positive samples.
Next, we uniformly sample some patches from set $\mathcal{D}^{\zeta,\beta}=\{\mathbf{x}||\zeta<||\mathbf{l}_t(\mathbf{x})-\mathbf{l}_t(\hat{\mathbf{x}}_t)||<\beta\}$,
and label them as negative samples.
After representing these samples with BMRs, we obtain a set of training data $\mathcal{D}^t=\{(\mathbf{b}(\mathbf{x}_t^i),y_t^i)\}_{i=1}^{n_t}$, where $y_t^i\in\{+1,-1\}$ is the class label and $n_t$ is the number of samples.
The cost function at frame $t$ is defined as the negative log-likelihood for logistic regression,
\begin{equation}
\ell_t(\mathbf{w}) = \frac{1}{n_t}\sum_{i=1}^{n_t}\log (1+\exp(-y_t^i\mathbf{w}^\top \mathbf{b}(\mathbf{x}_t^i))),
\label{eq:loglike}
\end{equation}
where $\mathbf{w}$ is the classifier parameter vector, and the corresponding classifier is denoted as
\begin{equation}
f(\mathbf{x}) = \frac{1}{1+\exp(-\mathbf{w}^\top \mathbf{b}(\mathbf{x}))}.
\label{eq:lrfunc}
\end{equation}

We use a gradient descent method to minimize $\ell_t(\mathbf{w})$ by iterating
\begin{equation}
\mathbf{w}\leftarrow \mathbf{w}-\frac{\partial \ell_t(\mathbf{w})}{\partial \mathbf{w}},
\label{eq:iterw}
\end{equation}
where $\frac{\partial \ell_t(\mathbf{w})}{\partial \mathbf{w}}\triangleq-\frac{1}{n_t}\sum_{i=1}^{n_t}\mathbf{b}(\mathbf{x}_t^i)\frac{y_t^i\exp(-y_t^i\mathbf{w}^\top \mathbf{b}(\mathbf{x}_t^i))}{1+\exp(-y_t^i\mathbf{w}^\top \mathbf{b}(\mathbf{x}_t^i))}$.
In this work, we use the parameter $\mathbf{w}_{t-1}$ obtained at frame $t-1$ to initialize $\mathbf{w}$ in (\ref{eq:iterw}) and iterate 20 times for updates.

\begin{algorithm}[t]
\caption{BMR-based Tracking}
\begin{algorithmic}\label{alg:BMRTrack}
\STATE \textbf{Input:} Target state $\mathbf{\hat{s}}_{t-1}$, classifier parameter vector $\mathbf{w}_t$;
\begin{enumerate}\setlength{\itemsep}{-\itemsep}
\item Sample $n_p$ candidate particles $\{\mathbf{s}_t^i\}_{i=1}^{n_p}$ with the motion  model $p(\mathbf{s}_t^i|\mathbf{\hat{s}}_{t-1})$ in (\ref{eq:particleapproximate});
\item For each particle $\mathbf{s}_t^i$, extract the corresponding image patch $\mathbf{x}_t^i$, and compute the BMR $\mathbf{b}(\mathbf{x}_t^i)$ by Algorithm~\ref{alg:BMR}, and compute
the corresponding observation model $p(\mathbf{o}_t|\mathbf{s}_t^i)$ by (\ref{eq:observation});
\item Estimate the optimal state $\mathbf{\hat{s}}_t$ by (\ref{eq:particleapproximate}), and
obtain the corresponding image patch $\hat{\mathbf{x}}_t$;
\item \textbf{if} $f(\hat{\mathbf{x}}_t)<\rho$
\item  \quad Update $\mathbf{w}_t$ by iterating (\ref{eq:iterw}) until convergence, and set $\mathbf{w}_{t+1}\leftarrow \mathbf{w}_t$;
\item \textbf{else}
\item \quad $\mathbf{w}_{t+1}\leftarrow\mathbf{w}_{t}$;
\item \textbf{end if}
\end{enumerate}
\STATE \textbf{Output:} Target state $\mathbf{\hat{s}}_t$ and classifier parameter vector $\mathbf{w}_{t+1}$.
\end{algorithmic}
\end{algorithm}
\subsection{Proposed Tracking Algorithm}
We estimate the target states sequentially within a particle filter framework.
Given the observation set $\mathcal{O}_t=\{\mathbf{o}_i\}_{i=1}^t$ up to frame $t$, the target sate $\mathbf{s}_t$ is obtained by maximizing the posteriori probability
\begin{equation}
\hat{\mathbf{s}}_t=\arg\max_{\mathbf{s}_t} \{p(\mathbf{s}_t|\mathcal{O}_t)\propto p(\mathbf{o}_t|\mathbf{s}_t)p(\mathbf{s}_t|\mathcal{O}_{t-1})\},
\label{eq:particle}
\end{equation}
where $p(\mathbf{s}_t|\mathcal{O}_{t-1})\triangleq\int p(\mathbf{s}_t|\mathbf{s}_{t-1})p(\mathbf{s}_{t-1}|\mathcal{O}_{t-1})d\mathbf{s}_{t-1}$, $\mathbf{s}_t=[x_t,y_t,s_t]$ is the target state with translations $x_t$ and $y_t$, and scale $s_t$, $p(\mathbf{s}_t|\mathbf{s}_{t-1})$ is a dynamic model that describes the temporal correlation of the target states in two consecutive frames, and $p(\mathbf{o}_t|\mathbf{s}_t)$ is the observation model that estimates the likelihood of a state given an observation.
In the proposed algorithm, we assume that the target state parameters are independent
and modeled by three scalar Gaussian distributions between two consecutive frames, i.e., $p(\mathbf{s}_t|\mathbf{s}_{t-1})=\mathcal{N}(\mathbf{s}_t|\mathbf{s}_{t-1},\mathbf{\Sigma})$, where $\mathbf{\Sigma}=\mathrm{diag}(\sigma_x,\sigma_y,\sigma_s)$ is a diagonal covariance matrix whose elements are the standard deviations of the  target state parameters.
In visual tracking, the posterior probability $p(\mathbf{s}_t|\mathcal{O}_t)$ in (\ref{eq:particle}) is approximated by a finite set of particles $\{\mathbf{s}_{t}^i\}_{i=1}^{n_p}$ that are sampled with corresponding importance weights $\{\pi_t^i\}_{i=1}^{n_p}$, where $\pi_t^i\propto p(\mathbf{o}_t|\mathbf{s}_t^i)$.
Therefore, (\ref{eq:particle}) can be approximated as
\begin{equation}
\hat{\mathbf{s}}_t = \arg\max_{\{\mathbf{s}_t^i\}_{i=1}^{n_p}}p(\mathbf{o}_t|\mathbf{s}_t^i)p(\mathbf{s}_t^i|\hat{\mathbf{s}}_{t-1}).
\label{eq:particleapproximate}
\end{equation}
In our method, the observation model $p(\mathbf{o}_t|\mathbf{s}_t^i)$ is defined as
\begin{equation}
p(\mathbf{o}_t|\mathbf{s}_t^i) \propto f(\mathbf{x}_t^i),
\label{eq:observation}
\end{equation}
where $f(\mathbf{x}_t^i)$ is the logistic regression classifier defined by (\ref{eq:lrfunc}).

To adapt to target appearance variations while preserving the stable information that helps prevent the tracker from drifting to background,  we update the classifier parameters $\mathbf{w}$ in a conservative way.
We update $\mathbf{w}$ by (\ref{eq:iterw})
only when the confidence of the target falls below a threshold $\rho$.
This ensures that the target states always have high confidence scores and alleviate the problem of including noisy samples when updating classifier~\cite{wang2015understanding}.
The main steps of the proposed algorithm are summarized in Algorithm~\ref{alg:BMRTrack}.

\section{Experimental Results}
We first present implementation details of the proposed algorithm, and
discuss the dataset and metrics for performance evaluation.
Next, we analyze the empirical results using widely-adopted metrics.
We present ablation study to examine the effectiveness of each key component in the proposed BMR scheme.
Finally, we show and analyze some failure cases.

\subsection{Implementation Details}
All images are resized to a fixed size of $240\times 320$ pixels~\cite{wang2015understanding} for experiments and
each patch is resized to a canonical size of $32\times 32$ pixels.
In addition, each canonical patch is subsampled to a half size with $n^{col}=16$ for
color representations.
The HOG features are extracted from the canonical patches that supports both gray and color images, and the sizes of HOG feature maps are the same as $n^{hog}\times n^{hog}\times 31=4\times 4\times 31$ (as implemented in {\url{http:///github.com/pdollar/toolbox}}).

For grayscale videos, the original image patches are used to extract raw intensity and HOG features, and the feature dimension $d=4\times 4\times 31+16\times 16 = 752$.
For color videos, the image patches are transformed to the CIE LAB color space to extract raw color features, and the original RGB image patches are used to extract HOG features.
The corresponding total dimension $d=4\times 4\times 31+16\times 16\times 3 = 1264$.
The number of Boolean maps is set to $c=4$, and the total dimension of BMRs is $3d=2256$ for gray videos, and $3792$ for color videos, and the sampling step $\delta=1/c=0.25$.
The search radius for positive samples is set to $\alpha=3$.
The inner search radius for negative samples is set to $0.3\min(w,h)$, where $w$ and $h$ are the weight and height of the target, respectively, and the outer search radius $\beta=100$, where the search step is set to $5$, which generates a small subset of negative samples.
The target state parameter set for particle filter is set to $[\sigma_x,\sigma_y,\sigma_s]=[6, 6, 0.01]$,
and the number of particles is set to $n_p=400$.
The confidence threshold is set to $\rho=0.9$.
%
All parameter values are fixed for all sequences and the source code will be made available to the public. More results and videos are available at {\url{http://kaihuazhang.net/bmr/bmr.htm}}.

\subsection{Dataset and Evaluation Metrics}
For performance evaluation, we use the tracking benchmark dataset and code library~\cite{wu2013online} which includes 29 trackers and 50 fully-annotated videos.
In addition, we also add the corresponding results of 6 most recent trackers including DLT~\cite{wang2013learning}, DSST~\cite{danelljan2014accurate},
KCF~\cite{henriques2015high}, TGPR~\cite{gao2014transfer}, MEEM~\cite{zhang2014meem}, and HCF~\cite{ma2015hierarchical}.
For detailed analysis, the sequences are annotated with 11 attributes based on different challenging factors
including low resolution (LR), in-plane rotation (IPR), out-of-plane rotation
(OPR), scale variation (SV), occlusion (OCC), deformation
(DEF), background clutters (BC), illumination variation (IV),
motion blur (MB), fast motion (FM), and out-of-view (OV).

We quantitatively evaluate the trackers with success and precision plots~\cite{wu2013online}.
%
Given the tracked bounding box $B_T$ and the ground truth bounding box $B_G$, the overlap score is defined as
$score=\frac{\textrm{Area}(B_T\bigcap B_G)}{\textrm{Area}(B_T\bigcup B_G)}.$
%
Hence, $0\leq score \leq 1$ and a larger value of $score$ means a better performance of the evaluated tracker.
The success plot demonstrates the percentage of frames with $score>t$ through all threshold $t\in [0,1]$.
Furthermore, the area under curve (AUC) of each success plot serves as a measure to rank the evaluated trackers.
On the other hand, the precision plot shows the percentage of frames whose tracked locations are within a given threshold distance
(i.e., 20 pixels in~\cite{wu2013online}) to the ground truth.
Both success and precision plots are used in the one-pass evaluation (OPE), temporal robustness evaluation (TRE), and spatial robustness evaluation (SRE) where
OPE reports the average precision or success rate by running the trackers through a test sequence with initialization from the ground truth position, and
TRE as well as SRE measure a tracker$'$s robustness to initialization with temporal and spatial perturbations, respectively~\cite{wu2013online}.
We report the OPE, TRE, and SRE results.
For presentation clarity, we only present the top 10 algorithms in each plot.
\begin{figure*}[ht]
\begin{center}
 \includegraphics[width=.32\linewidth]{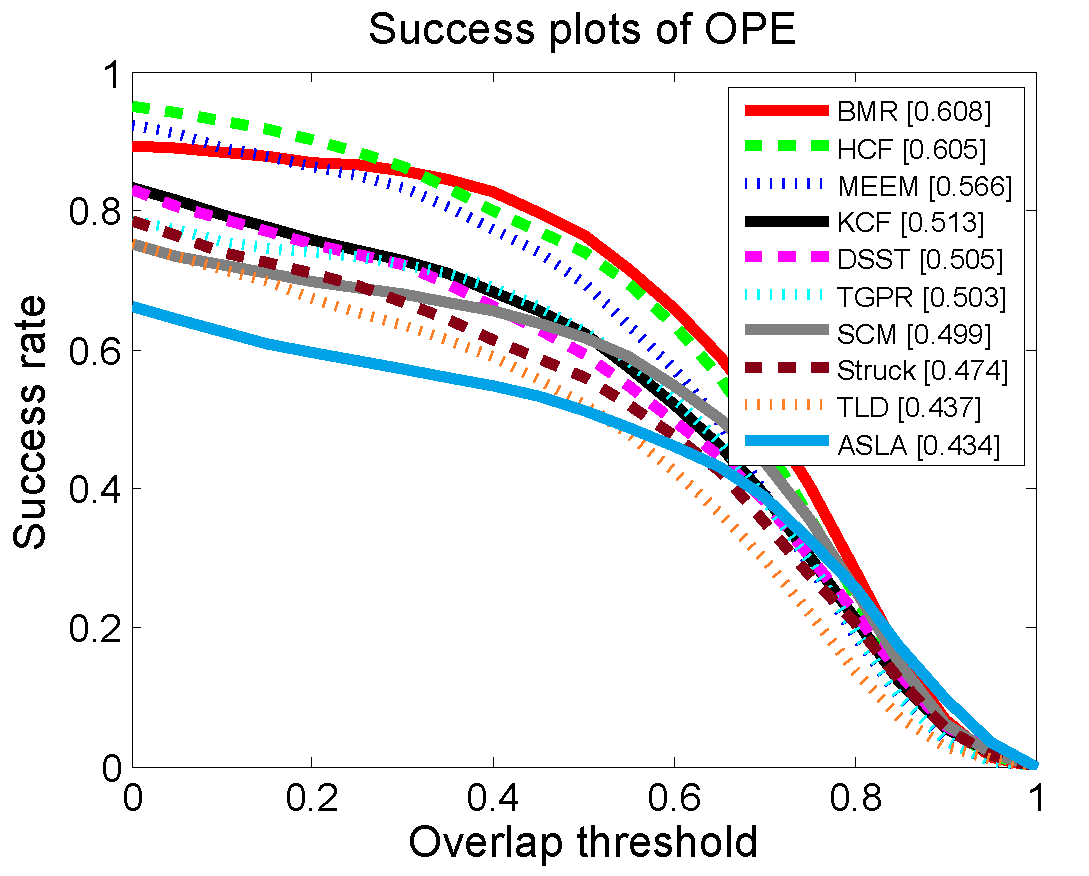}
 \includegraphics[width=.32\linewidth]{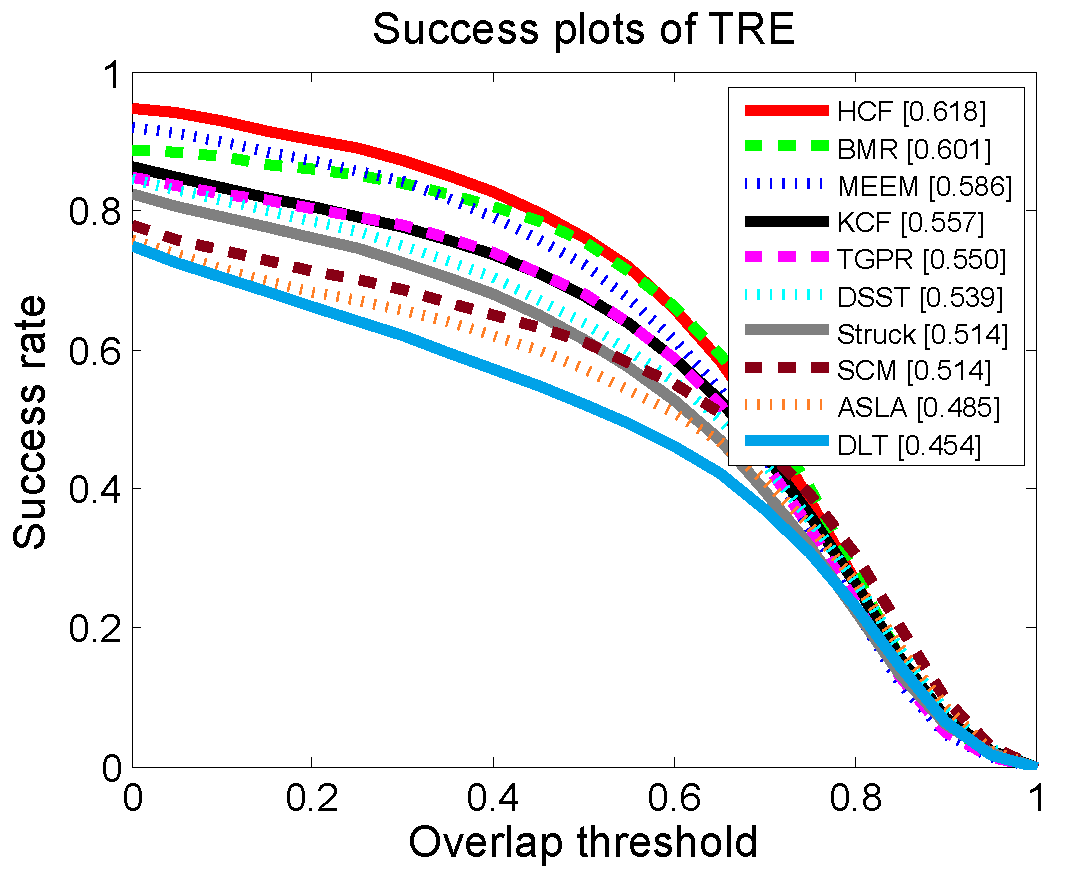}
 \includegraphics[width=.32\linewidth]{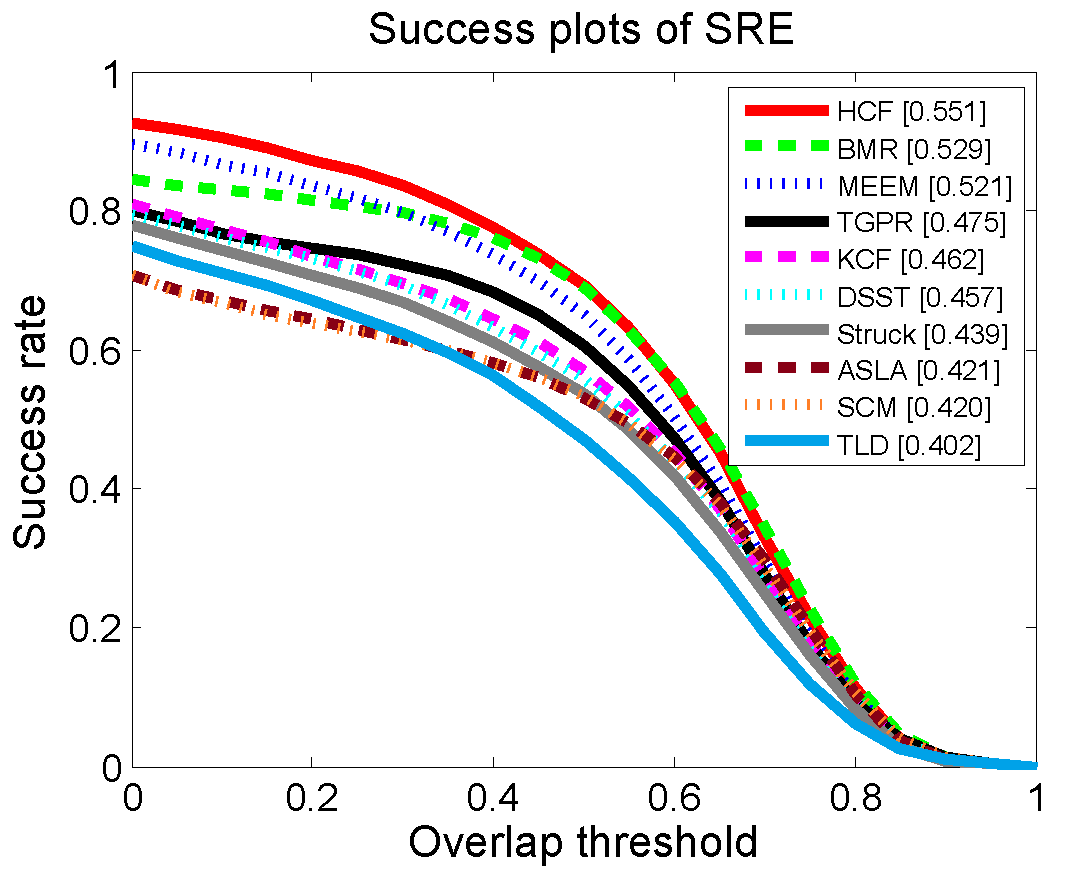}\\
  \includegraphics[width=.32\linewidth]{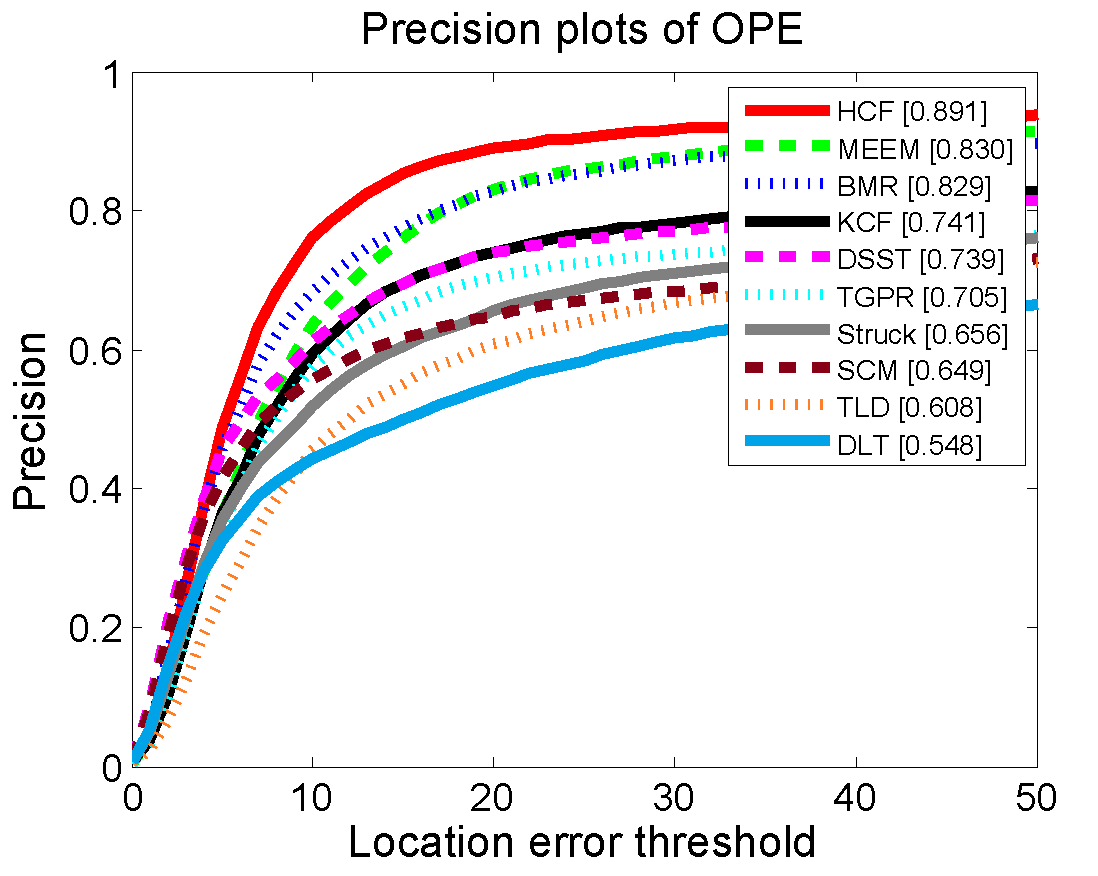}
 \includegraphics[width=.32\linewidth]{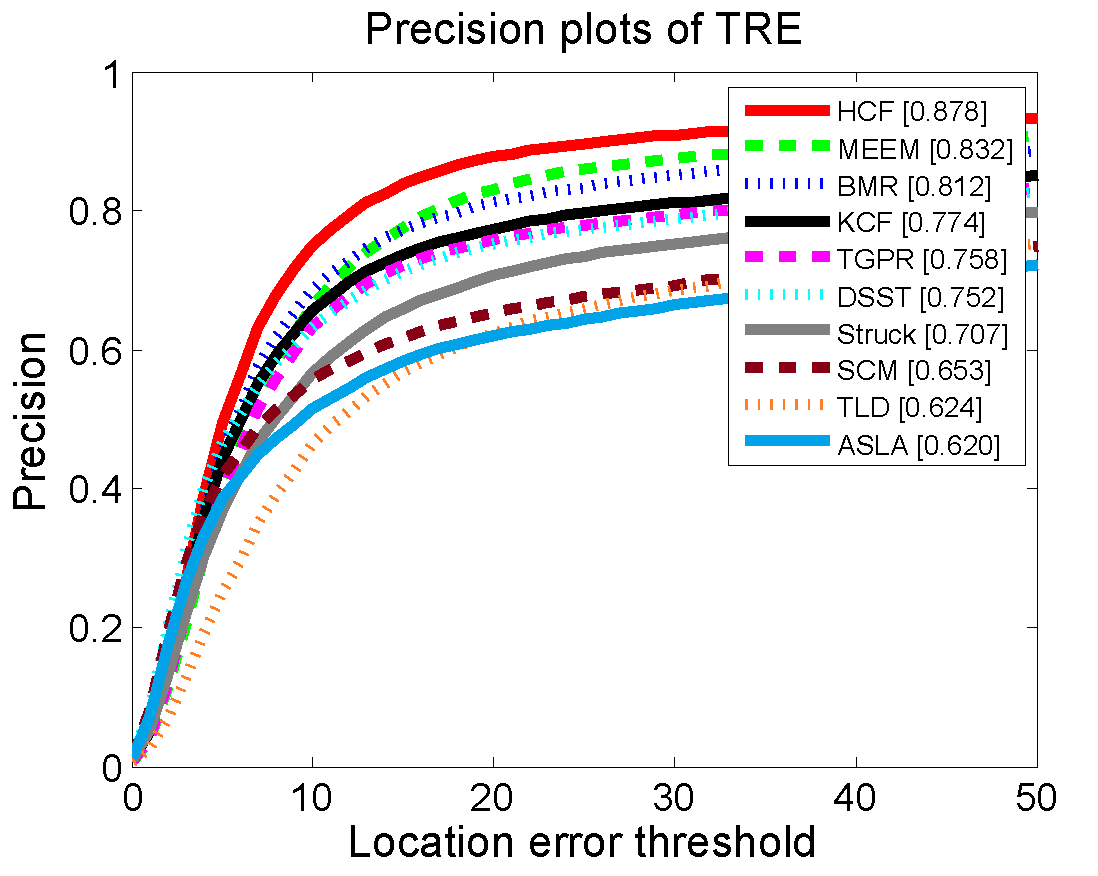}
 \includegraphics[width=.32\linewidth]{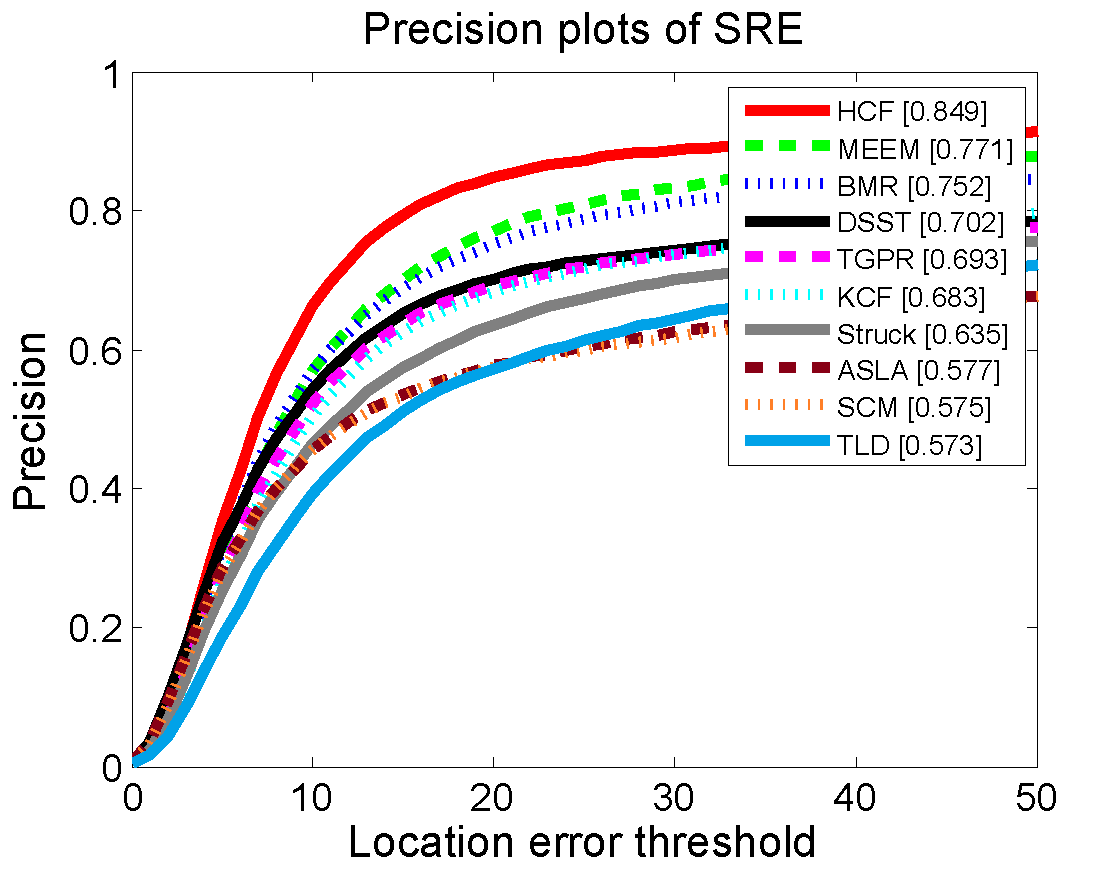}
\end{center}
\caption{Success and precision plots of OPE, TRE, and SRE by the top 10 trackers.
The trackers are ranked by the AUC scores (shown in the legends) when the success rates are used, or
precession cores at the threshold of 20 pixels. }
\label{fig:overallperformance}
\end{figure*}

\subsection{Empirical Results}
\textit{1) Overall Performance:} Figure~\ref{fig:overallperformance} shows overall performance of the top 10 trackers in terms of success and precision plots.
The BMR-based tracking algorithm ranks first on the success rate of all OPE,  and second based on TRE and SRE.
Furthermore, the BMR-based method ranks third based on the precision rates of OPE, TRE, and SRE.
Overall, the proposed BMR-based tracker performs favorably against the state-of-the-art methods in terms of all metrics except for
MEEM~\cite{zhang2014meem} and HCF~\cite{ma2015hierarchical}.
The MEEM tracker exploits a multi-expert restoration scheme to handle the drift problem, which combines a tracker and the historical
snapshots as experts.
In contrast, even using only a logistic regression classifier without using any restoration strategy, the proposed BMR-based method performs well against MEEM
in terms of most metrics (i.e., the success rates of the BMR-based method outperform the MEEM scheme while the precision rates of the BMR-based method
are comparable to the MEEM scheme),
which shows the effectiveness of the proposed representation scheme for visual tracking.
In addition, the HCF method is based on deep learning, which leverages complex hierarchical convolutional features learned off-line
from a large dataset and correlation filters for visual tracking.
Notwithstanding, the  proposed BMR-based algorithm performs comparably against HCF
in terms of success rates on all metrics.

%
%
\begin{table*}[ht]
\caption{Success score of OPE with 11 attributes. The number after
each attribute name is the number of sequences. The \textcolor{red}{red},  \textcolor{blue}{blue}
and \textcolor{green}{green} fonts indicate the best, second and third performance.
}
\scriptsize
\label{table:sucessScoreOPE}
{ \center\begin{tabular*}{\linewidth}
{@{\extracolsep{\fill}}|r||c|c|c|c|c|c|c|c|c|c|c|}\hline
Attribute            &BMR      &HCF~\cite{ma2015hierarchical}     &MEEM~\cite{zhang2014meem}      &KCF~\cite{henriques2015high}      &DSST~\cite{danelljan2014accurate}       &TGPR~\cite{gao2014transfer}     &SCM~\cite{zhong2012robust}     &Struck~\cite{Hare_ICCV_2011}   &TLD~\cite{Kalal_CVPR_2010} &ASLA~\cite{jia2012visual} &DLT~\cite{wang2013learning}    \\ \hline\hline
LR (4)    &\textcolor{blue}{0.409}           &\textcolor{red}{0.557}  &0.360        &0.310       &0.352    &\textcolor{green}{0.370}   &0.279  &0.372   &0.309  &0.157 &0.256   \\ \hline

IPR (31) &\textcolor{blue}{0.557}           &\textcolor{red}{0.582}  &\textcolor{green}{0.535}        &0.497       &0.532     &0.479  &0.458  &0.444   &0.416  &0.425 &0.383   \\ \hline

OPR (39) &\textcolor{red}{0.590}       &\textcolor{blue}{0.587}  &\textcolor{green}{0.558}        &0.496       &0.491     &0.485  &0.470  &0.432   &0.420  &0.422 &0.393 \\ \hline

SV (28)  &\textcolor{red}{0.586} &\textcolor{blue}{0.531} &\textcolor{green}{0.498}  &0.427       &0.451     &0.418     &0.518     &0.425         &0.421 &0.452 &0.458    \\ \hline

OCC (29) &\textcolor{red}{0.615} &\textcolor{blue}{0.606}  &\textcolor{green}{0.552} &0.513    &0.480     &0.484 &0.487        &0.413    &0.402 &0.376 &0.384  \\ \hline

DEF (19) &\textcolor{blue}{0.594} &\textcolor{red}{0.626}  &\textcolor{green}{0.560} &0.533    &0.474     &0.510 &0.448        &0.393    &0.378 &0.372 &0.330 \\ \hline

BC (21) &\textcolor{green}{0.555}   &\textcolor{red}{0.623}  &\textcolor{blue}{0.569}  &0.533   &0.492      &0.522 &0.450     &0.458     &0.345   &0.408 &0.327\\ \hline

IV (25) &\textcolor{blue}{0.551} &\textcolor{red}{0.560} &\textcolor{green}{0.533} &0.494   &0.506      &0.484     &0.473     &0.428         &0.399 &0.429 &0.392  \\ \hline

MB (12) &\textcolor{blue}{0.559} &\textcolor{red}{0.616} &\textcolor{green}{0.541} &0.499   &0.458     &0.434     &0.298     &0.433         &0.404       &0.258 &0.329 \\ \hline

FM (17) &\textcolor{blue}{0.559}    &\textcolor{red}{0.578} &\textcolor{green}{0.553} &0.461   &0.433     &0.396     &0.296     &0.462         &0.417  &0.247 &0.353  \\ \hline

OV (6) &\textcolor{red}{0.616} &\textcolor{green}{0.575} &\textcolor{blue}{0.606} &0.550    &0.490     &0.442     &0.361     &0.459         &0.457        &0.312 &0.409  \\ \hline

\end{tabular*}
}
\end{table*}
\begin{table*}[!ht]
\caption{Precision scores of OPE with 11 attributes. The number after
each attribute name is the number of sequences. The \textcolor{red}{red},  \textcolor{blue}{blue}
and \textcolor{green}{green} fonts indicate the best, second and third performance.
}
\label{table:PrecisionScoreOPE}
\scriptsize
{ \center\begin{tabular*}{\linewidth}
{@{\extracolsep{\fill}}|r||c|c|c|c|c|c|c|c|c|c|c|}\hline
Attribute                     &BMR      &HCF~\cite{ma2015hierarchical}     &MEEM~\cite{zhang2014meem}      &KCF~\cite{henriques2015high}      &DSST~\cite{danelljan2014accurate}       &TGPR~\cite{gao2014transfer}     &SCM~\cite{zhong2012robust}     &Struck~\cite{Hare_ICCV_2011}   &TLD~\cite{Kalal_CVPR_2010} &ASLA~\cite{jia2012visual} &DLT~\cite{wang2013learning}      \\ \hline\hline
LR (4)       &0.517     &\textcolor{red}{0.897}   &0.490             &0.379       &0.534    &\textcolor{green}{0.538}   &0.305   &\textcolor{blue}{0.545}   &0.349 &0.156 &0.303   \\ \hline

IPR (31)    &0.776     &\textcolor{red}{0.868}   &\textcolor{blue}{0.800}             &0.725       &\textcolor{green}{0.780}    &0.675   &0.597   &0.617   &0.584 &0.511 &0.510   \\ \hline

OPR (39) &\textcolor{green}{0.819}     &\textcolor{red}{0.869}   &\textcolor{blue}{0.840}             &0.730       &0.732    &0.678   &0.618   &0.597   &0.596 &0.518 &0.527 \\ \hline

SV (28)      &\textcolor{blue}{0.803}     &\textcolor{red}{0.880}   &\textcolor{green}{0.785}             &0.680       &0.740   &0.620   &0.672   &0.639 &0.606 &0.552 &0.606    \\ \hline

OCC (29)    &\textcolor{blue}{0.846}     &\textcolor{red}{0.877}   &\textcolor{green}{0.799}             &0.749       &0.725    &0.675   &0.640   &0.564   &0.563 &0.460 &0.495  \\ \hline

DEF (19)   &\textcolor{green}{0.802}     &\textcolor{red}{0.881}   &\textcolor{blue}{0.846}             &0.741       &0.657    &0.691   &0.586   &0.521   &0.512 &0.445 &0.512 \\ \hline

BC (21)   &0.742     &\textcolor{red}{0.885}   &\textcolor{blue}{0.797}             &\textcolor{green}{0.752}       &0.691    &0.717   &0.578   &0.585   &0.428 &0.496 &0.440\\ \hline

IV (25) &\textcolor{green}{0.742}    &\textcolor{red}{0.844}   &\textcolor{blue}{0.766}             &0.729       &0.741    &0.671   &0.594   &0.558   &0.537 &0.517 &0.492  \\ \hline

MB (12)     &\textcolor{blue}{0.755}     &\textcolor{red}{0.844}   &\textcolor{green}{0.715}             &0.650      &0.603    &0.537   &0.339   &0.551   &0.518 &0.278 &0.427 \\ \hline

FM (17)          &\textcolor{blue}{0.758}     &\textcolor{red}{0.790}  &\textcolor{green}{0.742}              &0.602       &0.562    &0.493   &0.333   &0.604   &0.551 &0.253 &0.435  \\ \hline

OV (6)          &\textcolor{red}{0.773}     &\textcolor{green}{0.695}  &\textcolor{blue}{0.727}              &0.649       &0.533    &0.505   &0.429   &0.539   &0.576 &0.333 &0.505  \\ \hline

\end{tabular*}
}
\end{table*}
\begin{table*}[ht]
\caption{Success scores of TRE with 11 attributes.
The number after
each attribute name is the number of sequences. The \textcolor{red}{red},  \textcolor{blue}{blue}
and \textcolor{green}{green} fonts indicate the best, second and third performance.
}
\scriptsize
\label{table:sucessScoreTRE}
{ \center\begin{tabular*}{\linewidth}
{@{\extracolsep{\fill}}|r||c|c|c|c|c|c|c|c|c|c|c|}\hline
Attribute            &BMR      &HCF~\cite{ma2015hierarchical}     &MEEM~\cite{zhang2014meem}      &KCF~\cite{henriques2015high}      &DSST~\cite{danelljan2014accurate}       &TGPR~\cite{gao2014transfer}     &SCM~\cite{zhong2012robust}     &Struck~\cite{Hare_ICCV_2011}   &TLD~\cite{Kalal_CVPR_2010} &ASLA~\cite{jia2012visual} &DLT~\cite{wang2013learning}    \\ \hline\hline
LR (4)    &\textcolor{green}{0.444}           &\textcolor{red}{0.520}  &0.424        &0.382       &0.403    &0.443   &0.304  &\textcolor{blue}{0.456}   &0.299  &0.278 &0.324   \\ \hline

IPR (31) &\textcolor{blue}{0.562}           &\textcolor{red}{0.591}  &\textcolor{green}{0.558}        &0.520       &0.515     &0.514  &0.453  &0.473   &0.406  &0.451 &0.423   \\ \hline

OPR (39) &\textcolor{blue}{0.578}       &\textcolor{red}{0.595}  &\textcolor{green}{0.572}        &0.531       &0.507     &0.523  &0.480  &0.477   &0.425  &0.465 &0.428 \\ \hline

SV (28)  &\textcolor{red}{0.564} &\textcolor{blue}{0.544} &\textcolor{green}{0.517}  &0.488       &0.473     &0.468     &0.496     &0.446         &0.418 &0.487 &0.448    \\ \hline

OCC (29) &\textcolor{blue}{0.585} &\textcolor{red}{0.610}  &\textcolor{green}{0.566} &0.547    &0.519     &0.520 &0.502        &0.462    &0.426 &0.444 &0.426  \\ \hline

DEF (19) &\textcolor{green}{0.599} &\textcolor{red}{0.651}  &\textcolor{blue}{0.611} &0.571    &0.548     &0.577 &0.515        &0.500    &0.425 &0.466 &0.399 \\ \hline

BC (21) &\textcolor{green}{0.575}   &\textcolor{red}{0.631}  &\textcolor{blue}{0.577}  &0.565   &0.518      &0.530 &0.469     &0.478     &0.372   &0.445 &0.366\\ \hline

IV (25) &\textcolor{green}{0.555} &\textcolor{red}{0.597} &\textcolor{blue}{0.564} &0.528   &0.529      &0.518     &0.475     &0.486         &0.402 &0.468 &0.427  \\ \hline

MB (12) &\textcolor{green}{0.537} &\textcolor{red}{0.594} &\textcolor{blue}{0.553} &0.493   &0.472     &0.483     &0.290     &0.485         &0.388       &0.296 &0.349 \\ \hline

FM (17) &\textcolor{green}{0.516}    &\textcolor{red}{0.560} &\textcolor{blue}{0.542} &0.456   &0.429     &0.461     &0.282     &0.464         &0.392  &0.285 &0.350  \\ \hline

OV (6) &\textcolor{red}{0.593} &\textcolor{green}{0.557} &\textcolor{blue}{0.581} &0.539    &0.505     &0.440     &0.344     &0.417         &0.434        &0.325 &0.403  \\ \hline

\end{tabular*}
}
\end{table*}
\begin{table*}[!ht]
\caption{Precision scores of TRE with 11 attributes.
The number after
each attribute name is the number of sequences. The \textcolor{red}{red},  \textcolor{blue}{blue}
and \textcolor{green}{green} fonts indicate the best, second and third performance.
}
\label{table:PrecisionScoreTRE}
\scriptsize
{ \center\begin{tabular*}{\linewidth}
{@{\extracolsep{\fill}}|r||c|c|c|c|c|c|c|c|c|c|c|}\hline
Attribute                     &BMR      &HCF~\cite{ma2015hierarchical}     &MEEM~\cite{zhang2014meem}      &KCF~\cite{henriques2015high}      &DSST~\cite{danelljan2014accurate}       &TGPR~\cite{gao2014transfer}     &SCM~\cite{zhong2012robust}     &Struck~\cite{Hare_ICCV_2011}   &TLD~\cite{Kalal_CVPR_2010} &ASLA~\cite{jia2012visual} &DLT~\cite{wang2013learning}      \\ \hline\hline
LR (4)        &0.581     &\textcolor{red}{0.750}   &\textcolor{green}{0.589}             &0.501       &0.574    &\textcolor{blue}{0.602}   &0.350   &0.628   &0.376 &0.325 &0.391   \\ \hline

IPR (31)     &\textcolor{green}{0.767}     &\textcolor{red}{0.851}   &\textcolor{blue}{0.802}             &0.728       &0.725    &0.716   &0.581   &0.650   &0.569 &0.582 &0.572   \\ \hline

OPR (39) &\textcolor{green}{0.789}     &\textcolor{red}{0.859}   &\textcolor{blue}{0.826}             &0.749       &0.719    &0.728   &0.617   &0.660   &0.597 &0.605 &0.584 \\ \hline

SV (28)       &\textcolor{green}{0.769}     &\textcolor{red}{0.840}   &\textcolor{blue}{0.787}             &0.727       &0.717   &0.676   &0.633   &0.652 &0.600 &0.634 &0.594    \\ \hline

OCC (29)    &\textcolor{blue}{0.791}     &\textcolor{red}{0.854}   &\textcolor{green}{0.788}             &0.758       &0.726    &0.705   &0.633   &0.631   &0.579 &0.560 &0.550  \\ \hline

DEF (19)   &\textcolor{green}{0.798}     &\textcolor{red}{0.889}   &\textcolor{blue}{0.854}             &0.757       &0.723    &0.765   &0.635   &0.655   &0.571 &0.571 &0.556 \\ \hline

BC (21)   &0.772     &\textcolor{red}{0.874}   &\textcolor{blue}{0.793}             &\textcolor{green}{0.776}       &0.697    &0.721   &0.600   &0.622   &0.488 &0.575 &0.517\\ \hline

IV (25) &\textcolor{green}{0.747}    &\textcolor{red}{0.851}   &\textcolor{blue}{0.792}             &0.729       &0.727    &0.693   &0.585   &0.643   &0.543 &0.584 &0.572  \\ \hline

MB (12)     &\textcolor{green}{0.720}     &\textcolor{red}{0.785}   &\textcolor{blue}{0.724}             &0.626      &0.597    &0.607   &0.323   &0.617   &0.491 &0.332 &0.450 \\ \hline

FM (17)          &\textcolor{green}{0.681}     &\textcolor{red}{0.738}  &\textcolor{blue}{0.710}              &0.578      &0.532    &0.582   &0.302   &0.580   &0.487 &0.305 &0.432  \\ \hline

OV (6)          &\textcolor{red}{0.719}     &\textcolor{blue}{0.692}  &\textcolor{blue}{0.692}              &\textcolor{green}{0.643}       &0.587    &0.514   &0.371  &0.484   &0.485 &0.339 &0.470  \\ \hline

\end{tabular*}
}
\end{table*}
\begin{table*}[!ht]
\caption{Success scores of SRE with 11 attributes.
The number after
each attribute name is the number of sequences. The \textcolor{red}{red},  \textcolor{blue}{blue}
and \textcolor{green}{green} fonts indicate the best, second and third performance.
}
\scriptsize
\label{table:sucessScoreSRE}
{ \center\begin{tabular*}{\linewidth}
{@{\extracolsep{\fill}}|r||c|c|c|c|c|c|c|c|c|c|c|}\hline
Attribute            &BMR      &HCF~\cite{ma2015hierarchical}     &MEEM~\cite{zhang2014meem}      &KCF~\cite{henriques2015high}      &DSST~\cite{danelljan2014accurate}       &TGPR~\cite{gao2014transfer}     &SCM~\cite{zhong2012robust}     &Struck~\cite{Hare_ICCV_2011}   &TLD~\cite{Kalal_CVPR_2010} &ASLA~\cite{jia2012visual} &DLT~\cite{wang2013learning}    \\ \hline\hline
LR (4)   &\textcolor{green}{0.352}           &\textcolor{red}{0.488}  &\textcolor{blue}{0.374}        &0.289       &0.326    &0.332   &0.254  &0.360   &0.305  &0.213 &0.243   \\ \hline

IPR (31) &\textcolor{green}{0.487}           &\textcolor{red}{0.537}  &\textcolor{blue}{0.494}        &0.450       &0.460    &0.438  &0.399  &0.410   &0.380  &0.405 &0.357   \\ \hline

OPR (39) &\textcolor{green}{0.510}       &\textcolor{red}{0.536}  &\textcolor{blue}{0.514}        &0.445       &0.439     &0.455  &0.396  &0.409   &0.387  &0.404 &0.368 \\ \hline

SV (28)  &\textcolor{red}{0.524} &\textcolor{blue}{0.492} &\textcolor{green}{0.463}  &0.401       &0.413     &0.396     &0.438     &0.395         &0.384 &0.440 &0.402    \\ \hline

OCC (29) &\textcolor{blue}{0.524} &\textcolor{red}{0.543}  &\textcolor{green}{0.510} &0.445    &0.434     &0.449 &0.398        &0.405    &0.384 &0.381 &0.354  \\ \hline

DEF (19) &\textcolor{green}{0.492} &\textcolor{red}{0.566}  &\textcolor{blue}{0.516} &0.469    &0.434     &0.504 &0.358        &0.398    &0.357 &0.386 &0.322 \\ \hline

BC (21) &\textcolor{green}{0.500}   &\textcolor{red}{0.569}  &\textcolor{blue}{0.517}  &0.483   &0.451      &0.483 &0.387     &0.408     &0.334   &0.410 &0.303\\ \hline

IV (25) &\textcolor{green}{0.486} &\textcolor{red}{0.516} &\textcolor{blue}{0.490} &0.442   &0.446      &0.438     &0.389     &0.396         &0.350 &0.405 &0.347  \\ \hline

MB (12) &\textcolor{green}{0.503} &\textcolor{red}{0.565} &\textcolor{blue}{0.513} &0.425   &0.389     &0.420     &0.266     &0.451         &0.385       &0.256 &0.312 \\ \hline

FM (17) &\textcolor{green}{0.504}    &\textcolor{red}{0.534} &\textcolor{blue}{0.518} &0.415   &0.384     &0.412     &0.269     &0.464         &0.392  &0.285 &0.350  \\ \hline

OV (6) &\textcolor{red}{0.578} &\textcolor{green}{0.526} &\textcolor{blue}{0.575} &0.455    &0.426     &0.391     &0.335     &0.421         &0.407        &0.316 &0.314  \\ \hline

\end{tabular*}
}
\end{table*}
\begin{table*}[!ht]
\caption{Precision scores of SRE with 11 attributes.
The number after
each attribute name is the number of sequences. The \textcolor{red}{red},  \textcolor{blue}{blue}
and \textcolor{green}{green} fonts indicate the best, second and third performance.
}
\label{table:PrecisionScoreSRE}
\scriptsize
{ \center\begin{tabular*}{\linewidth}
{@{\extracolsep{\fill}}|r||c|c|c|c|c|c|c|c|c|c|c|}\hline
Attribute                     &BMR      &HCF~\cite{ma2015hierarchical}     &MEEM~\cite{zhang2014meem}      &KCF~\cite{henriques2015high}      &DSST~\cite{danelljan2014accurate}       &TGPR~\cite{gao2014transfer}     &SCM~\cite{zhong2012robust}     &Struck~\cite{Hare_ICCV_2011}   &TLD~\cite{Kalal_CVPR_2010} &ASLA~\cite{jia2012visual} &DLT~\cite{wang2013learning}      \\ \hline\hline
LR (4)       &0.476     &\textcolor{red}{0.818}   &\textcolor{green}{0.511}             &0.377       &\textcolor{blue}{0.543}    &0.501   &0.305   &0.504   &0.363 &0.263 &0.299   \\ \hline

IPR (31)    &\textcolor{green}{0.704}     &\textcolor{red}{0.839}   &\textcolor{blue}{0.752}             &0.667       &\textcolor{green}{0.704}    &0.648   &0.546   &0.592   &0.554 &0.556 &0.503   \\ \hline

OPR (39)     &\textcolor{green}{0.732}     &\textcolor{red}{0.828}   &\textcolor{blue}{0.774}             &0.666       &0.680    &0.669   &0.547   &0.595   &0.560 &0.560 &0.525 \\ \hline

SV (28)      &\textcolor{blue}{0.752}     &\textcolor{red}{0.832}   &\textcolor{green}{0.732}             &0.632       &0.696  &0.599   &0.598   &0.607 &0.558 &0.601 &0.562    \\ \hline

OCC (29)     &\textcolor{blue}{0.735}     &\textcolor{red}{0.815}   &\textcolor{green}{0.730}             &0.662       &0.671    &0.649   &0.540   &0.568   &0.516 &0.514 &0.483  \\ \hline

DEF (19)     &0.684     &\textcolor{red}{0.835}   &\textcolor{blue}{0.757}             &0.677       &0.630   &\textcolor{green}{0.715}   &0.475   &0.547   &0.505 &0.516 &0.467\\ \hline

BC (21)    &\textcolor{green}{0.702}     &\textcolor{red}{0.851}   &\textcolor{blue}{0.734}             &0.693       &0.655    &0.698   &0.521   &0.555   &0.451 &0.555 &0.439\\ \hline

IV (25)     &0.677    &\textcolor{red}{0.809}   &\textcolor{blue}{0.707}             &0.652       &\textcolor{green}{0.681}    &0.630   &0.509   &0.556   &0.480 &0.544 &0.472  \\ \hline

MB (12)     &\textcolor{green}{0.686}     &\textcolor{red}{0.807}   &\textcolor{blue}{0.691}             &0.567      &0.532    &0.561   &0.309   &0.587   &0.521 &0.310 &0.388 \\ \hline

FM (17)     &\textcolor{green}{0.685}     &\textcolor{red}{0.748}  &\textcolor{blue}{0.694}              &0.545      &0.505    &0.544   &0.308   &0.577   &0.496 &0.291 &0.397  \\ \hline

OV (6)    &\textcolor{red}{0.719}     &\textcolor{green}{0.644}  &\textcolor{blue}{0.690}              &0.533       &0.504    &0.451   &0.386  &0.455   &0.463 &0.355 &0.360  \\ \hline

\end{tabular*}
}
\end{table*}
\textit{2) Attribute-based Performance:} To demonstrate the strength and weakness of BMR, we further evaluate the 35 trackers on videos with 11 attributes categorized by~\cite{wu2013online}.
\begin{figure*}[ht]
\begin{center}
 \includegraphics[width=1\linewidth]{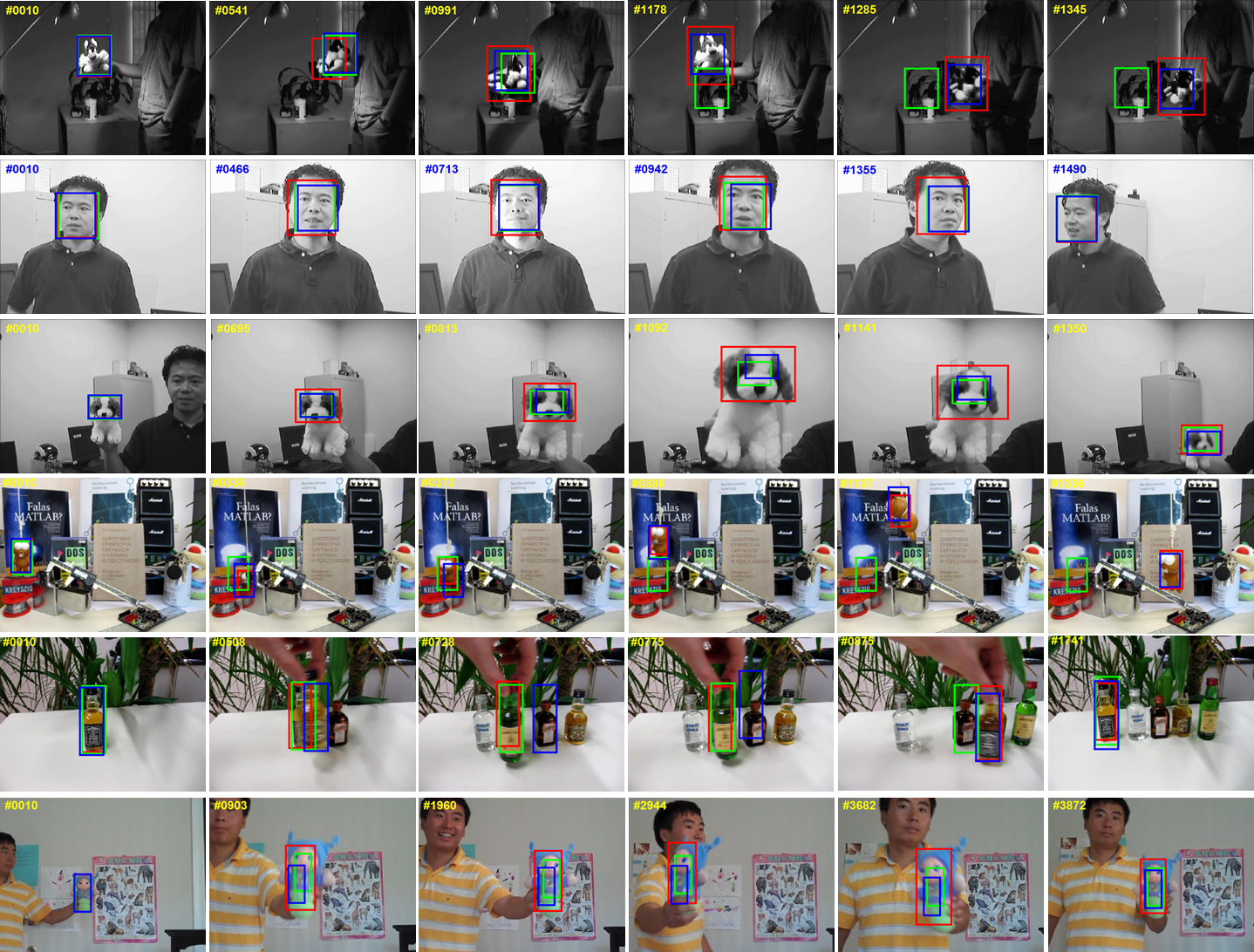}\\
 \includegraphics[width=0.2\linewidth]{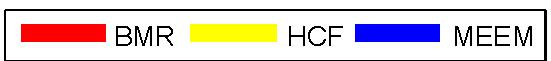}
 \end{center}
   \caption{Screenshots sampled results from six long sequences \textit{sylvester}, \textit{mhyang}, \textit{dog1}, \textit{lemming}, \textit{liquor}, and
\textit{doll}. }
\label{fig:longsequence}
\end{figure*}

Table~\ref{table:sucessScoreOPE} and \ref{table:PrecisionScoreOPE} summarize the results of success and precision scores of OPE with different attributes.
Among them, the BMR-based method ranks within top 3 with most attributes.
Specifically, with the success rate of OPE, the BMR-based method ranks first on 4 out of 11 attributes while second on 6 out of 11 attributes.
In the sequences with the BC attribute, the BMR-based method ranks third and its score is  close to the MEEM scheme that ranks second (0.555 vs. 0.569).
For the precision scores of OPE, the BMR-based method ranks second on 4 out of 11 attributes and third on 3 out of 11 attributes.
In the sequences with the OV attribute, the BMR-based tracker ranks first,
and for the videos with the IPR and BC attributes, the proposed tracking algorithm ranks fourth with comparable performance to the third-rank DSST and KCF methods.

Table~\ref{table:sucessScoreTRE} and \ref{table:PrecisionScoreTRE} show the results of TRE with different attributes.
The BMR-based method ranks within top 3 with most attributes.
In terms of success rates, the BMR-based method ranks first on 2 attributes, second on 3 attributes and third on 6 attributes.
In terms of precision rates, the BMR-based tracker ranks third on 7 attributes, and first and second on the OV and OCC attributes, respectively.
Furthermore, for other attributes such as LR and BC, the BMR-based tracking algorithm
ranks fourth but it scores are close to the results of MEEM and KCF that rank third
(0.581 vs. 0.598, and 0.772 vs. 0.776).

Table~\ref{table:sucessScoreSRE} and \ref{table:PrecisionScoreSRE} show the results of SRE with different attributes.
In terms of success rates, the rankings of the BMR-based method are similar to those
based on TRE except for the IPR and OPR attributes.
Among them, the BMR-based tracker ranks third based on SRE and second based on TRE.
Furthermore, although the MEEM method ranks higher than the BMR-based tracker
in most attributes,  the differences of the scores are within $1\%$.
In terms of precision rates, the BMR-based algorithm
ranks within top 3 with most attributes except for the LR, DEF, and IV attributes.

The AUC score of success rate measures the overall performance of each tracking method
\cite{wu2013online}.
Figure~\ref{fig:overallperformance} shows that
the BMR-based method achieves better results in terms of success rates than that
precision rates in terms of all metrics (OPE, SRE, TRE) and attributes.
The tracking performance can be attributed to two factors.
First, the proposed method exploits a logistic regression classifier with explicit feature maps, which efficiently determines the nonlinear decision boundary through online training.
Second, the online classifier parameter update scheme in~(\ref{eq:iterw}) facilitates
recovering from tracking drift.
\begin{figure*}[tb]
\begin{center}
 \includegraphics[width=0.49\linewidth]{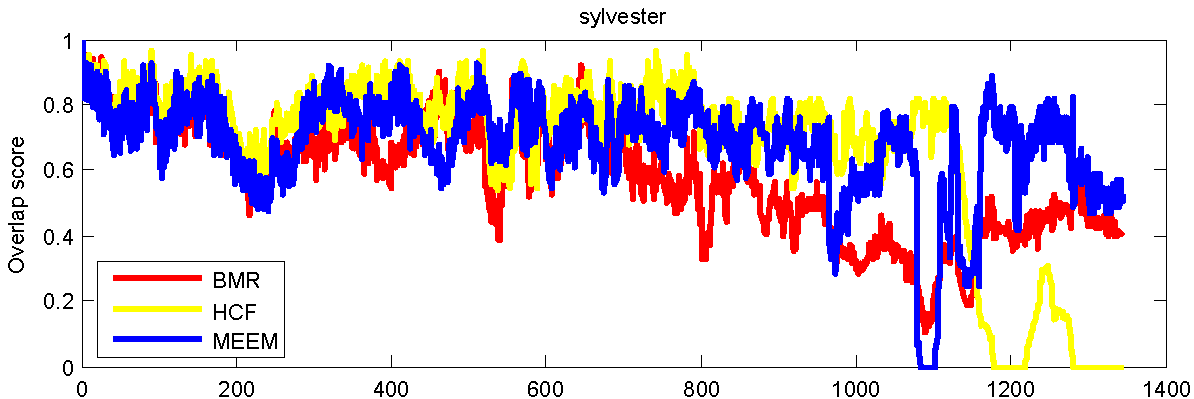}
 \includegraphics[width=0.49\linewidth]{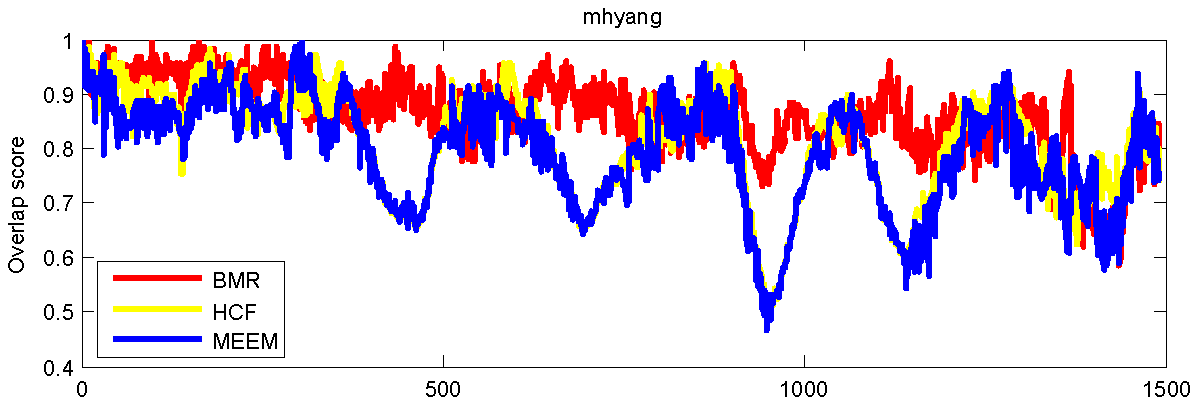}\\
 \includegraphics[width=0.49\linewidth]{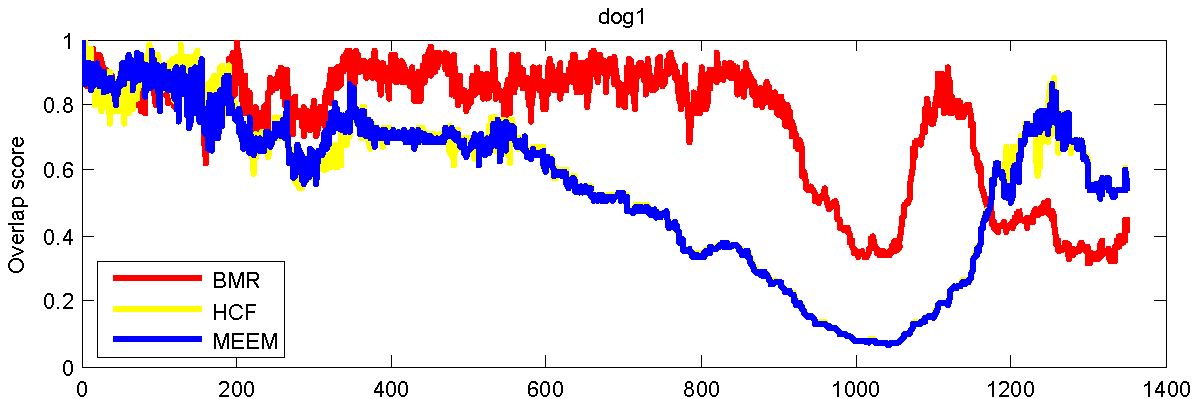}
 \includegraphics[width=0.49\linewidth]{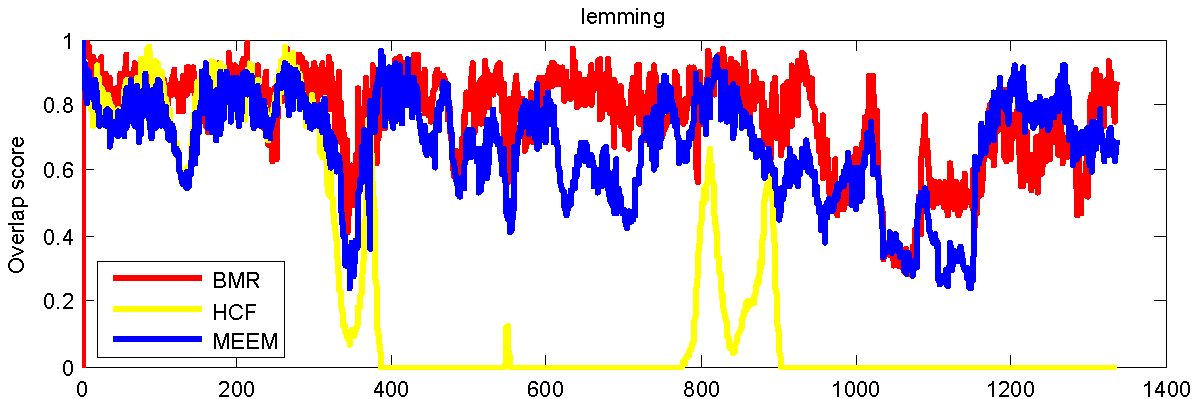}\\
 \includegraphics[width=0.49\linewidth]{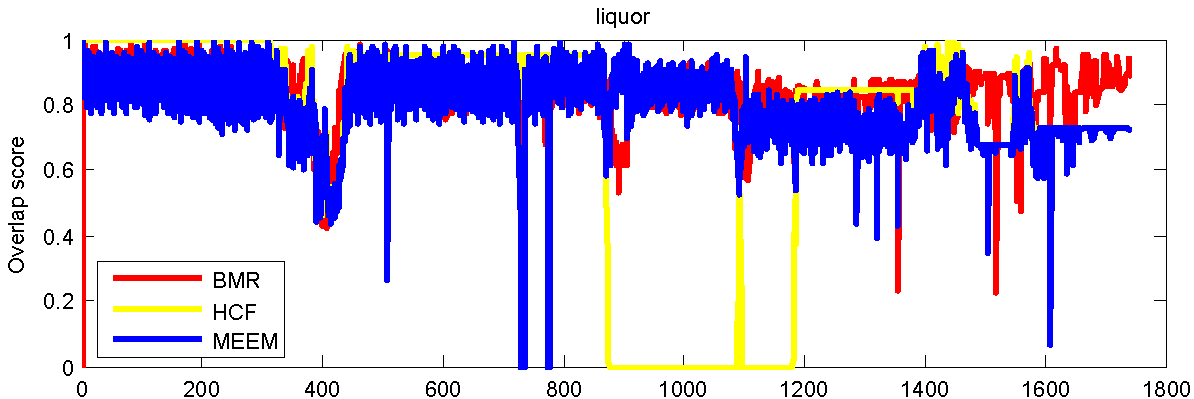}
  \includegraphics[width=0.49\linewidth]{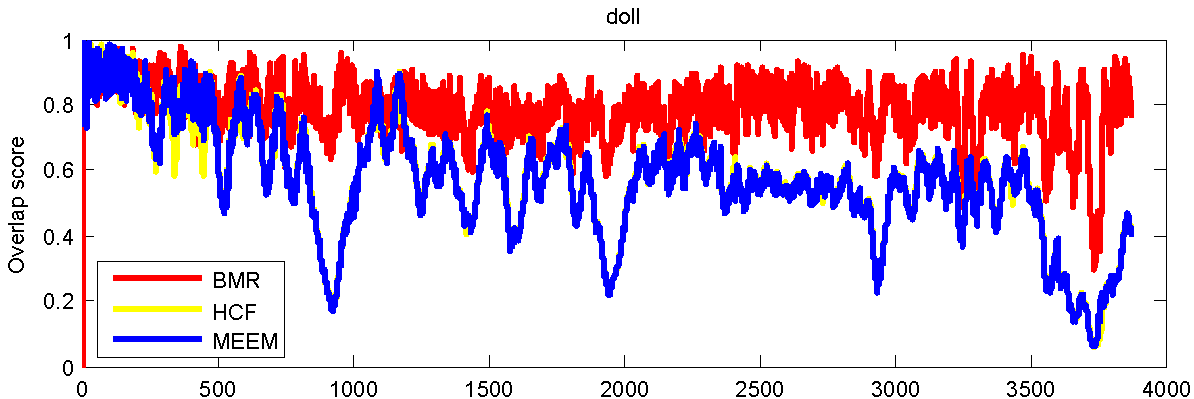}
  \end{center}
   \caption{Overlap score plots of six long sequences shown in Figure~\ref{fig:longsequence}.}
\label{fig:cle}
\end{figure*}

Figure~\ref{fig:longsequence} shows sampled tracking results from six long sequences (each with more than 1000 frames).
The total number of frames of these sequences is $11,134$ that accounts for about $38.4\%$ of the total number of frames (about $29,000$) in the benchmark, and hence the performance on these sequences plays an important role in performance evaluation.
For clear presentation, only the results of the top performing BMR, HCF, and MEEM
methods are  shown.
In all sequences, the BMR-based tracker is able to track the targets stably over almost all frames.
However, the HCF scheme drifts away from the target objects after a few frames in the \textit{sylvester} ($\#1178, \#1285, \#1345$) and \textit{lemming} ($\#386,\#1137,\#1336$) sequences.
The MEEM method drifts to background when severe occlusions happen in the \textit{liquor} sequence ($\#508,\#728,\#775$).
To further demonstrate the results over all frames clearly, Figure~\ref{fig:cle} shows
the plots in terms of overlap score of each frame.
Overall, the BMR-based tracker performs well  against the HCF and MEEM methods in
most frames of these sequences.
\begin{figure*}[tb]
\begin{center}
 \includegraphics[width=.32\linewidth]{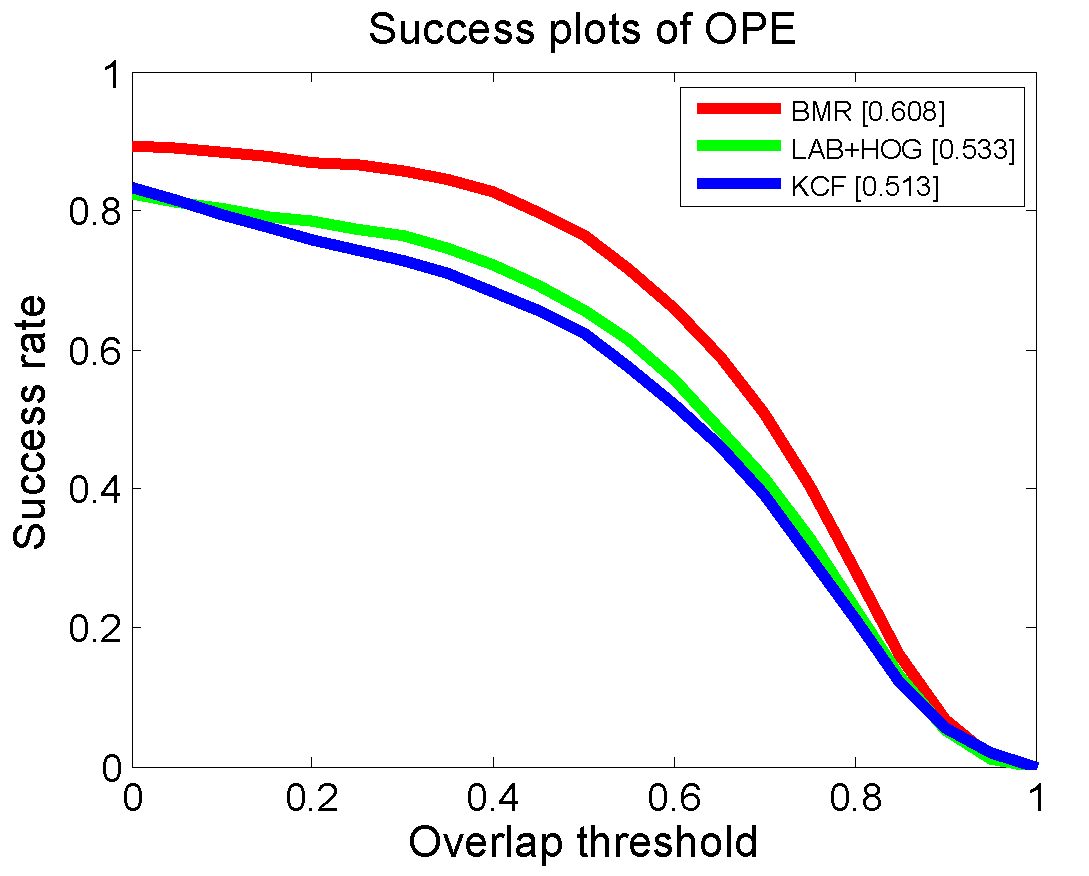}
 \includegraphics[width=.32\linewidth]{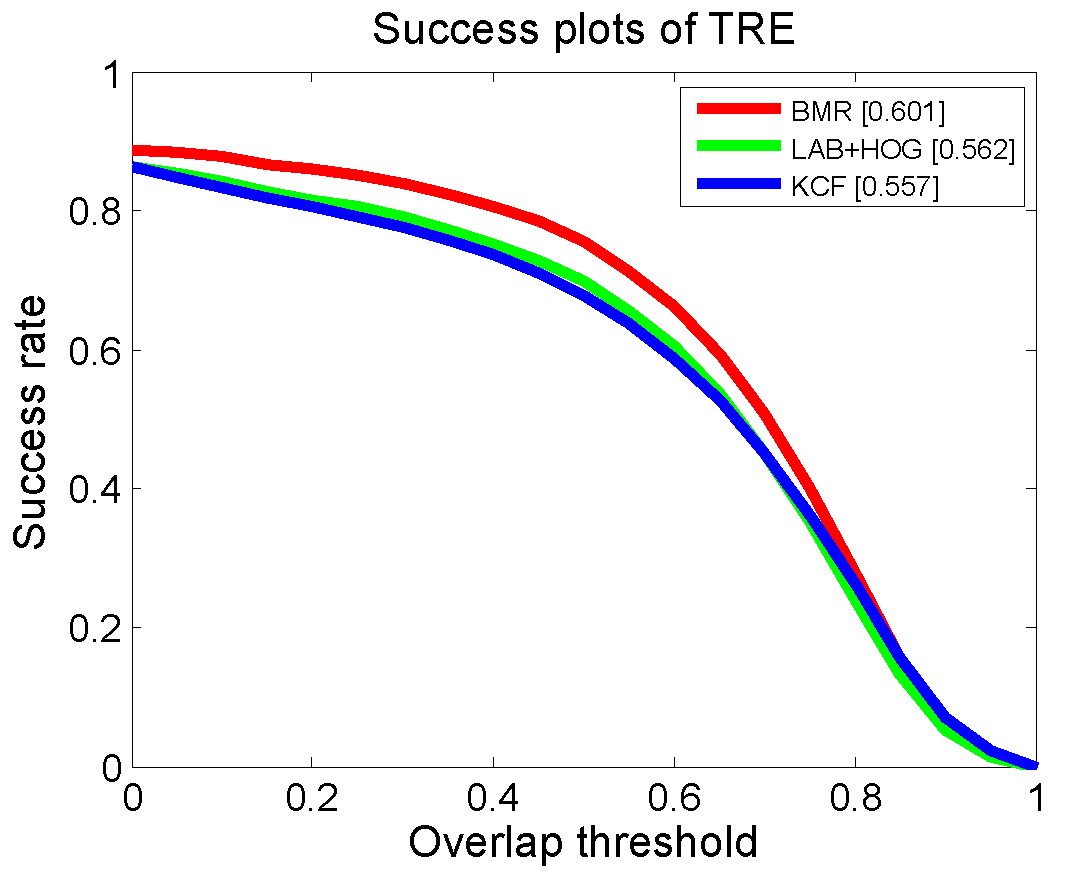}
 \includegraphics[width=.32\linewidth]{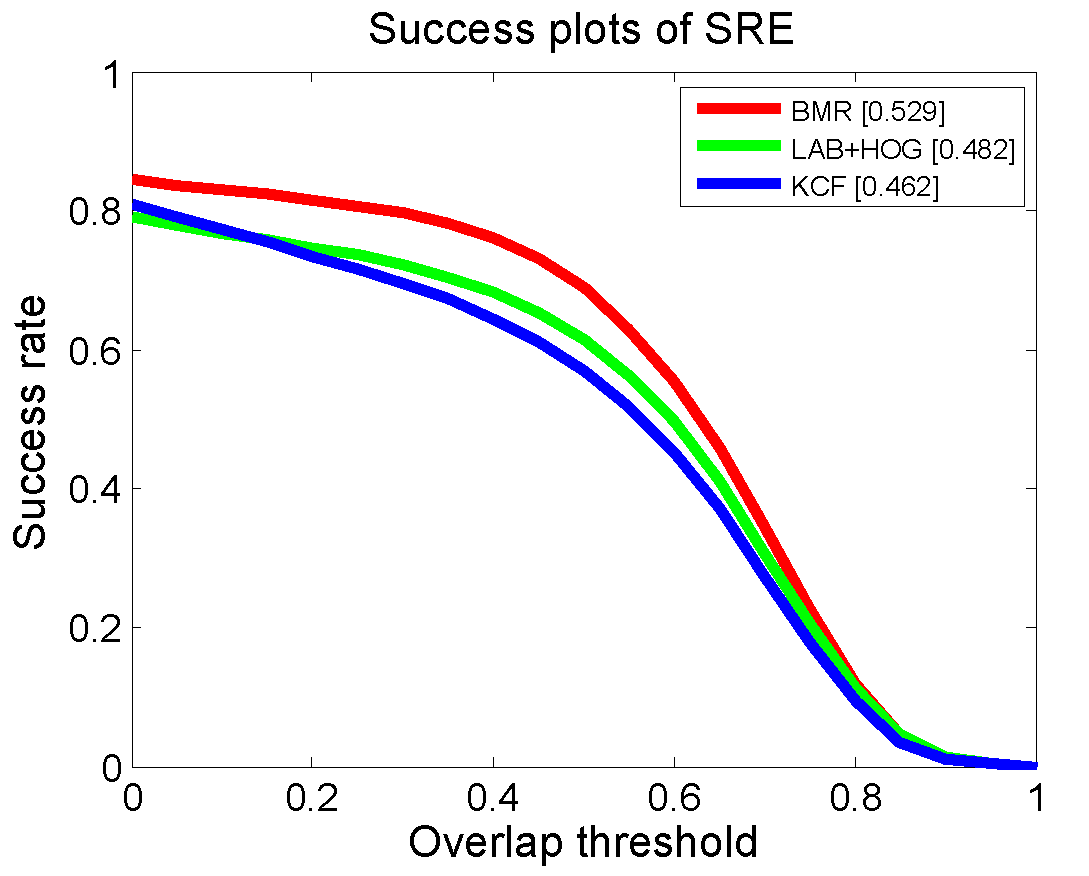}\\
  \includegraphics[width=.32\linewidth]{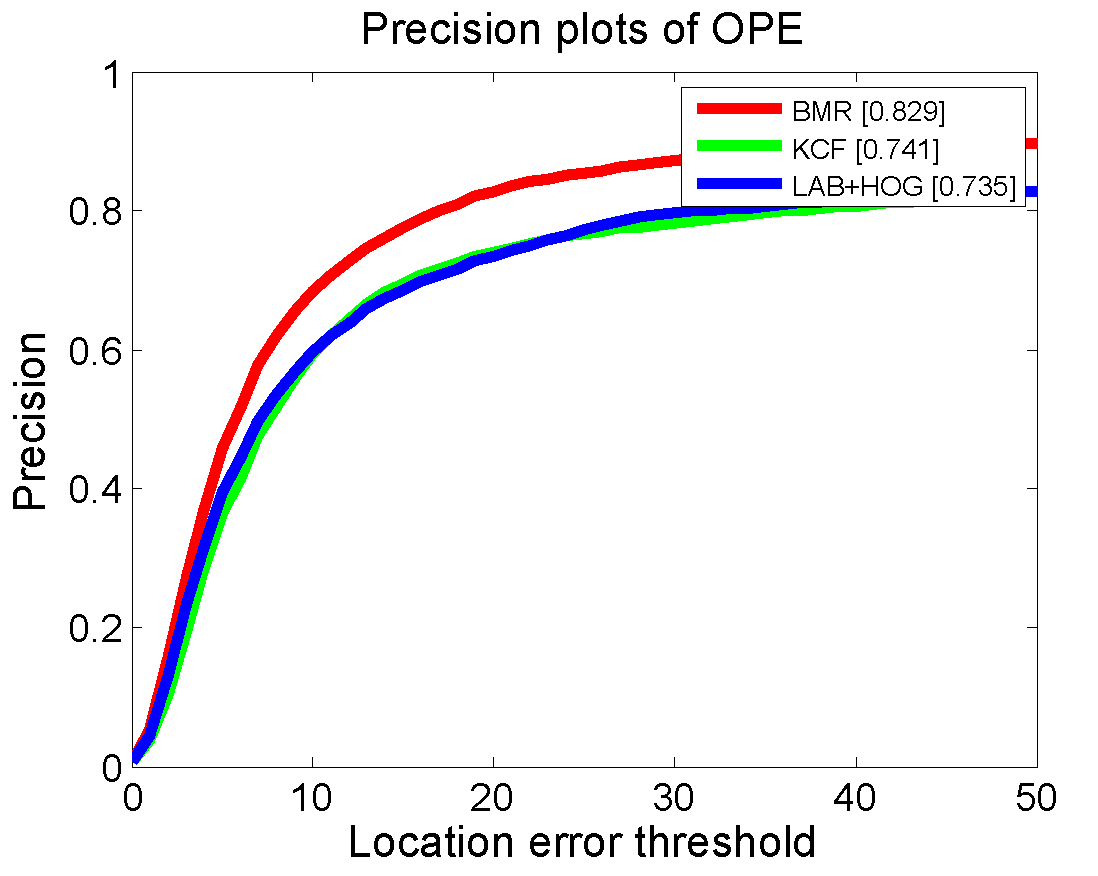}
 \includegraphics[width=.32\linewidth]{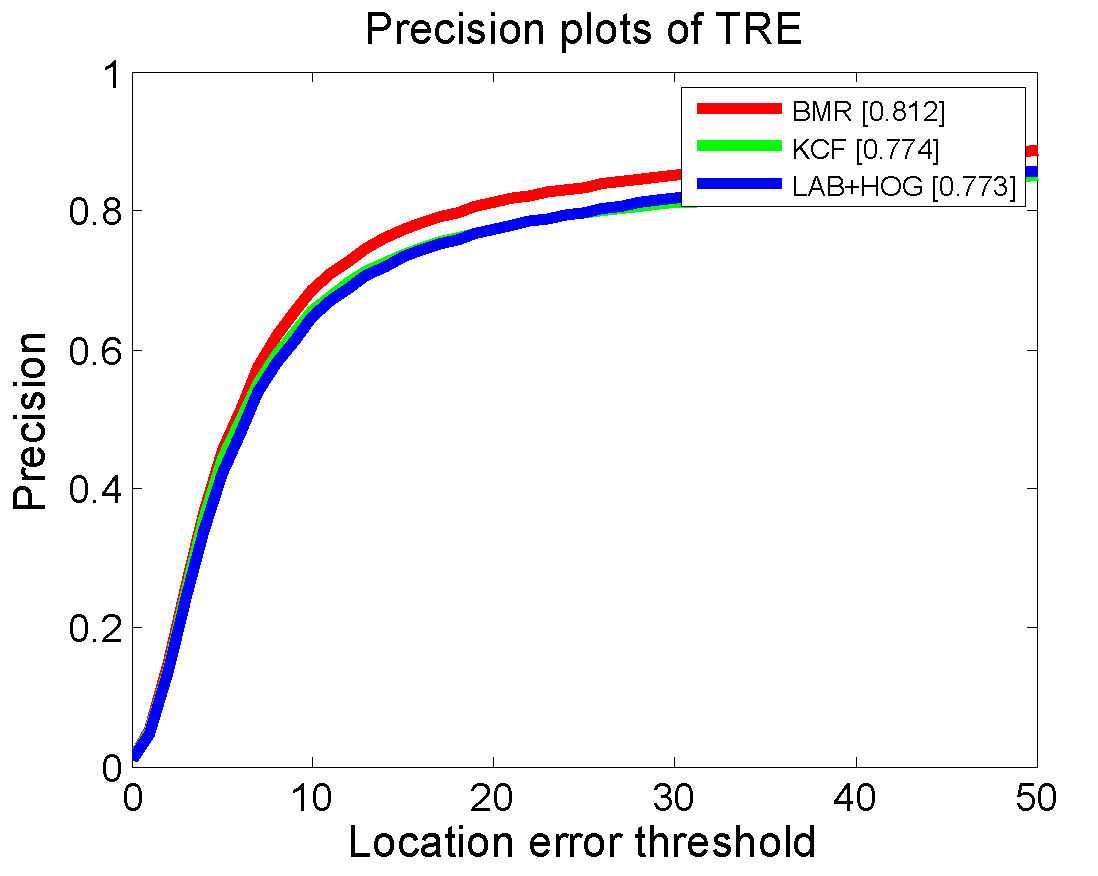}
 \includegraphics[width=.32\linewidth]{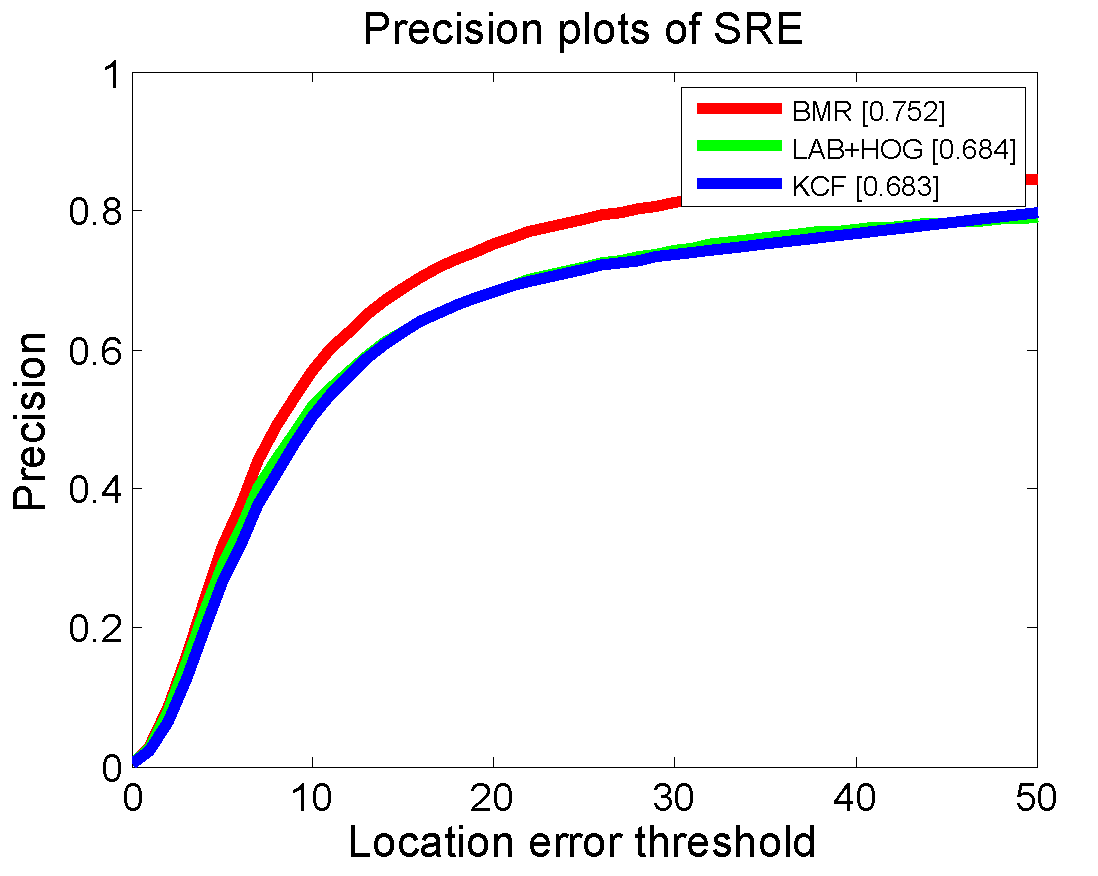}
\end{center}
   \caption{Success and precision plots of OPE, TRE, and SRE for BMR, BMR only with LAB+HOG representations, and KCF (KCF is used as a baseline for comparisons).}
\label{fig:analysis}
\end{figure*}

\subsection{Analysis of BMR}
To demonstrate the effectiveness of BMRs, we eliminate the component of Boolean maps in the proposed tracking algorithm and only leverage the LAB+HOG representations for visual tracking.
In addition, we use the KCF as a baseline as it adopts the HOG representations as the proposed tracking method.
Figure~\ref{fig:analysis} shows quantitative comparisons on the benchmark dataset.
Without using the proposed Boolean maps, the AUC score of success rate in OPE of the proposed method is reduced by $7.5\%$.
For TRE and SRE, the AUC scores of the proposed method are reduced by $3.9\%$ and $4.7\%$, respectively without the component of Boolean maps.
It is worth noticing that the proposed method, without using the Boolean maps, still outperforms KCF in terms of all metrics on success rates,
which shows the effectiveness of the LAB color features in BMR.
These experimental results show that the BMRs in the proposed method play a key role for robust visual tracking.

\begin{figure*}[tb]
\begin{center}
 \includegraphics[width=1\linewidth]{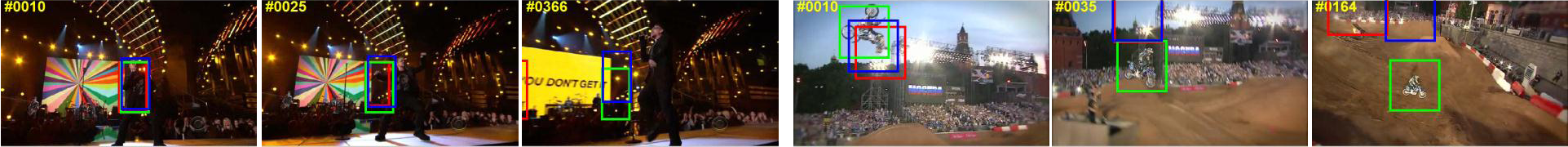}\\
 \includegraphics[width=.49\linewidth]{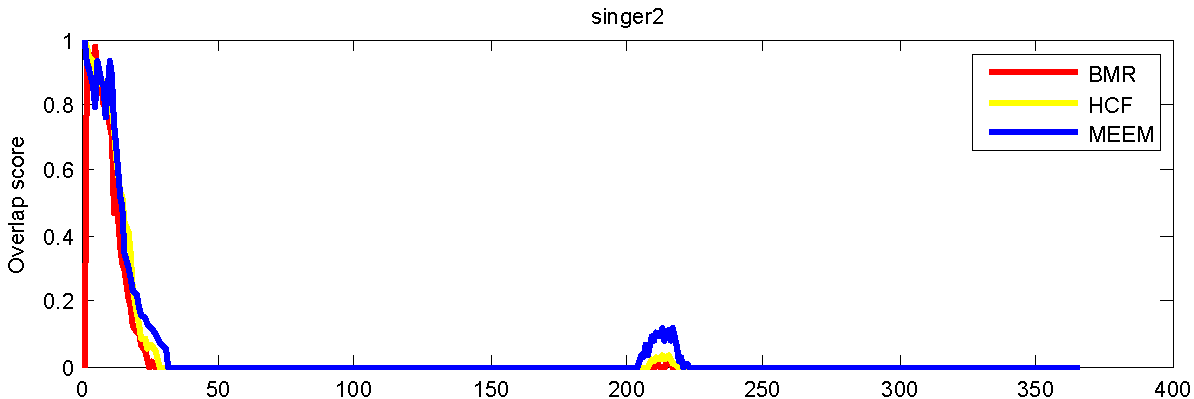}
 \includegraphics[width=.49\linewidth]{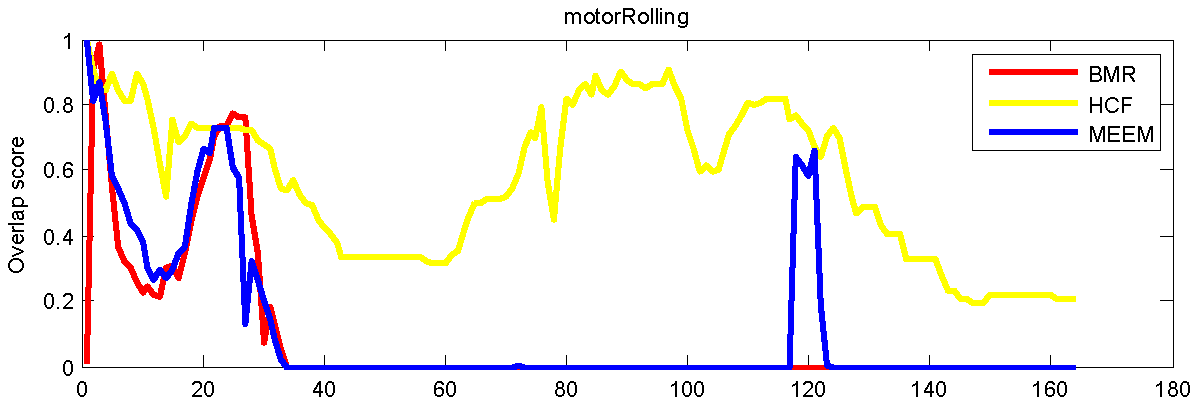}\\
 \end{center}
   \caption{Failure cases of the BMR-based tracker in the  \textit{singer2} and \textit{motorRolling} sequences. The results of HCF and MEEM are also illustrated.}
\label{fig:failure}
\end{figure*}

\subsection{Failure Cases}
Figure~\ref{fig:failure} shows failed results of the proposed BMR-based method in two sequences \textit{singer2} and \textit{motorRolling}.
In the \textit{singer2} sequence, the foreground object and background scene are similar due to the dim stage lighting at the beginning ($\#10,\#25$).
The HCF, MEEM and proposed methods all drift to the background.
Furthermore, as the targets in the \textit{motorRolling} sequence undergo from 360-degree in-plane rotation in early frames ($\#35$), the MEEM and proposed methods
do not adapt to drastic appearance variations well due to limited training samples.
In contrast, only the HCF tracker performs well in this sequence because it leverages dense sampling and high-dimensional convolutional features.
%

\section{Conclusions}
In this paper, we propose a Boolean map based representation which exploits the connectivity cues for visual tracking.
In the BMR scheme, the HOG and raw color feature maps are decomposed into a set of Boolean maps by uniformly thresholding the respective channels.
These Boolean maps are concatenated and normalized to form a robust representation, which approximates an explicit feature map of the intersection kernel.
A logistic regression classifier with the explicit feature map is trained in an online manner
that determines the nonlinear decision boundary for visual tracking.
Extensive evaluations on a large tracking benchmark dataset demonstrate the
proposed tracking algorithm performs favorably against the  state-of-the-art algorithms in terms of accuracy and robustness.

\bibliographystyle{ieeetr}
\bibliography{egbib}
\end{document}